\documentclass{article}

    \usepackage[final,nonatbib]{neurips_2022} 

\usepackage[utf8]{inputenc} 
\usepackage[T1]{fontenc}    
\usepackage{hyperref}       
\usepackage{url}            
\usepackage{booktabs}       
\usepackage{amsfonts}       
\usepackage{nicefrac}       
\usepackage{microtype}      
\usepackage{xcolor}         

\usepackage{comment}

\usepackage{hyperref}
\usepackage{multirow}
\usepackage{xcolor,pifont}
\newcommand*\colourcheck[1]{%
  \expandafter\newcommand\csname #1check\endcsname{\textcolor{#1}{\ding{52}}}%
}
\colourcheck{green}
\newcommand*\colourx[1]{%
  \expandafter\newcommand\csname #1x\endcsname{\textcolor{#1}{\ding{55}}}%
}
\colourx{red}
\usepackage{graphicx}
\usepackage{wrapfig}
\usepackage{makecell}

\usepackage{amsmath}
\usepackage{amssymb}
\usepackage{amsfonts}
\usepackage{bm}

\def\eqref#1{equation~\ref{#1}}

\def\1{\bm{1}}

\DeclareMathAlphabet{\mathsfit}{\encodingdefault}{\sfdefault}{m}{sl}
\SetMathAlphabet{\mathsfit}{bold}{\encodingdefault}{\sfdefault}{bx}{n}

\DeclareMathOperator*{\argmin}{arg\,min}

\title{Robust Feature-Level Adversaries\\are Interpretability Tools}

\author{
  Stephen Casper$^{*123}$, Max Nadeau$^{*234}$, Dylan Hadfield-Menell$^1$, Gabriel Kreiman$^{23}$ \\
  $^1$MIT CSAIL; $^2$ Boston Children’s Hospital, Harvard Medical School;\\
  $^3$Center for Brains, Minds, and
Machines; $^4$Harvard College, Harvard University\\
  \texttt{scasper@mit.edu} \hspace{0.75cm} \texttt{mnadeau@college.harvard.edu}\\
  $*$ Equal Contribution
}

\begin{document}

\maketitle

\begin{abstract}

The literature on adversarial attacks in computer vision typically focuses on pixel-level perturbations. These tend to be very difficult to interpret. Recent work that manipulates the latent representations of image generators to create ``feature-level'' adversarial perturbations gives us an opportunity to explore perceptible, interpretable adversarial attacks. We make three contributions. First, we observe that feature-level attacks provide useful classes of inputs for studying representations in models. Second, we show that these adversaries are uniquely versatile and highly robust. We demonstrate that they can be used to produce targeted, universal, disguised, physically-realizable, and black-box attacks at the ImageNet scale. Third, we show how these adversarial images can be used as a practical interpretability tool for identifying bugs in networks. We use these adversaries to make predictions about spurious associations between features and classes which we then test by designing ``copy/paste'' attacks in which one natural image is pasted into another to cause a targeted misclassification. Our results suggest that feature-level attacks are a promising approach for rigorous interpretability research. They support the design of tools to better understand what a model has learned and diagnose brittle feature associations.\footnote{\href{https://github.com/thestephencasper/feature_level_adv}{https://github.com/thestephencasper/feature\_level\_adv}.}

\end{abstract}

\section{Introduction}
\label{sec:introduction}

State-of-the-art neural networks are vulnerable to adversarial examples. 
Conventionally, adversarial inputs for visual classifiers take the form of small-norm perturbations to natural images \cite{szegedy2013intriguing, goodfellow2014explaining}.
These perturbations reliably cause confident misclassifications. 
However, to a human, they typically appear as random or mildly-textured noise. 
Consequently, it is difficult to interpret these attacks--they rarely generalize to produce human-comprehensible insights about the target network.
In other words, beyond the observation that such attacks are possible, it is hard to learn much about the underlying target network from these pixel-level perturbations.

In contrast, many real-world failures of biological vision are caused by perceptible, human-describable features. 
For instance, the ringlet butterfly's predators are stunned by adversarial "eyespots" on its wings (Appendix \ref{app:adversarial_features_in_nature}, Fig. \ref{fig:nature}). 
This falls outside the scope of conventional adversarial examples because the misclassification results from a feature-level change to an object/image. 
The adversarial eyespots are robust in the sense that the same attack works across a variety of different observers, backgrounds, and viewing conditions. 
Furthermore, because the attack relies on high-level features, it is easy for a human to describe it.

This work takes inspiration from the ringlet butterfly's eyespots and similar examples in which a model is fooled in the real world by an interpretable feature (e.g. \cite{ntsb2018collision}). 
Our goal is to design adversaries that reveal easily understandable weaknesses of the victim network. 
We focus on two desiderata for adversarial perturbations: attacks must be (1) interpretable (i.e. describable) to a human, and (2) robust so that interpretations generalize. 
We refer to these types of attacks as ``feature-level'' adversarial examples. 
Several previous works have created attacks by perturbing the latent representations of an image generator (e.g., \cite{hu2021naturalistic}), but thus far, approaches have been small in scale, limited in robustness, and not interpretability-driven (See Section \ref{sec:related_work}). 

We build on this prior work to propose an attack method that generates feature-level attacks against computer vision models. 
This method works on ImageNet scale models and creates robust, feature-level adversarial examples. 
We test three methods of introducing adversarial features into source images either by modifying the generator's latents and/or inserting a generated patch into natural images.
In contrast to previous works that have enforced the ``adversarialness'' of attacks only by inserting small features or restricting the distance between an adversary and a benign input, we also introduce methods that regularize the feature to be perceptible yet disguised to resemble something other than the target class.

\begin{figure*}[t!]
\centering
\includegraphics[width=\linewidth]{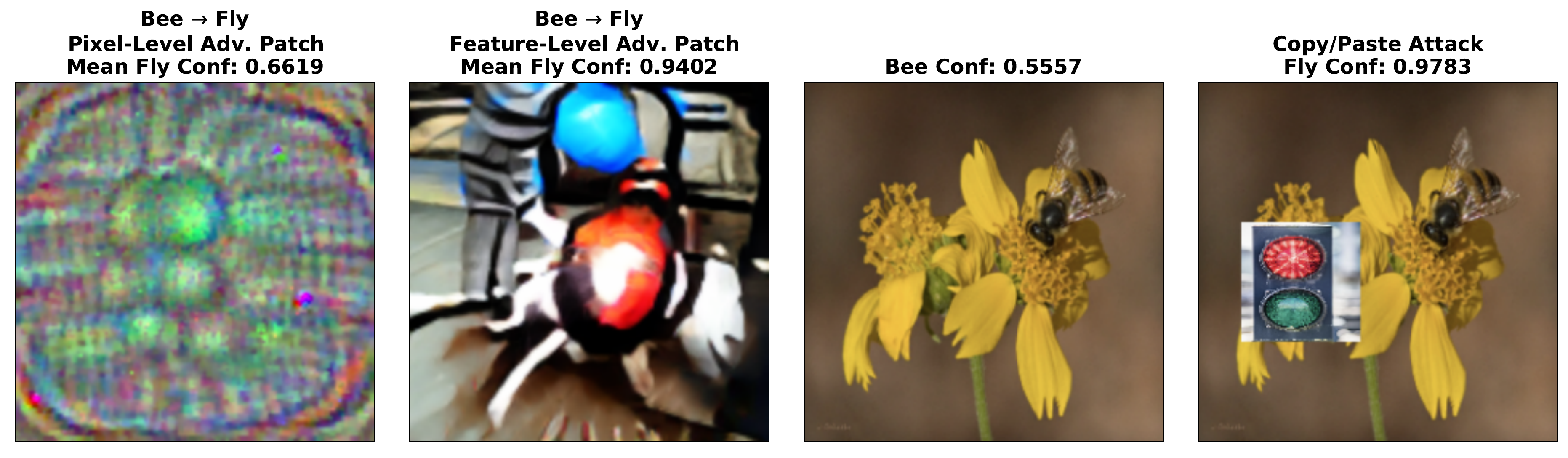}
\begin{tabular}{cccc}
    (a) & \hspace{1in} (b) & \hspace{1in} (c) & \hspace{1in} (d) 
\end{tabular}
\caption{Our feature-level adversaries are useful for interpreting deep networks (we used a ResNet50 \cite{he2016deep}). (a) A pixel-level adversarial patch trained to make images of bees misclassified as flies. (b) An analogous feature-level adversarial patch. (c) A correctly classified image of a bee. (d) A successful copy/paste attack whose design was guided by adversarial examples like the one in (b).
}
\label{fig:bee}
\end{figure*}

We show that our method produces robust attacks that provide actionable insights into a network's learned representations. 
Fig. \ref{fig:bee} demonstrates the interpretability benefits of this type of feature-level attack. 
It compares a conventional, pixel-level, adversarial patch, created using the method from \cite{brown2017adversarial}, with a feature-level attack using our method. 
While both attacks attempt to make a network misclassify a bee as a fly, the pixel-level attack exhibits high-frequency patterns and lacks visually coherent objects. 
On the other hand, the feature-level attack displays easily describable features: the colored circles.
We can validate this insight by considering the network performance when a picture of a traffic light is inserted into the image a bee. 
In this example, the image classification moves from a 55\% confidence that the image is a bee to a 97\% confidence that the image is of a fly. 
Section.~\ref{sec:interpretable_copy_paste_attacks} studies these types of ``copy/paste'' attacks more in-depth. 

Our contributions are threefold. 
\begin{enumerate}
    \item \textbf{Conceptual Insight:} We observe that robust feature-level adversaries can used to produce useful types of inputs for studying the representations of deep networks
    \item \textbf{Robust Attacks:} We introduce methods for generating feature-level adversaries that are uniquely versatile and able to produce targeted, universal, disguised, physically-realizable, and black-box attacks at the ImageNet scale. See Table \ref{tbl:comparisons}.
    \item \textbf{Interpretability:} We generalize from our adversarial examples to design copy/paste attacks, verifying that our adversaries help us understand the network well enough to exploit it.
\end{enumerate}
    
The following sections contain background, methods, experiments, and discussion. 
Appendix \ref{app:jargon_free_summary} has a high-level summary. 
Code is available at \href{https://github.com/thestephencasper/feature_level_adv}{https://github.com/thestephencasper/feature\_level\_adv}.

\section{Related Work}
\label{sec:related_work}

Here, we contextualize our approach with others related to improving on conventional adversarial examples \cite{szegedy2013intriguing, goodfellow2014explaining}. Table \ref{tbl:comparisons} summarizes capabilities.

\begin{table*}[t!]
\centering
\scriptsize
\begin{tabular}{|l|c|c|c|c|c|c|c|}
\hline

& \multirow{2}{*}{\textbf{Targeted}} & \multirow{2}{*}{\textbf{Universal}} & \multirow{2}{*}{\textbf{Disguised}} & \textbf{Physically-} & \textbf{Transferable/} & \multirow{2}{*}{\textbf{Copy/Paste}} & \textbf{ImageNet} \\
& & & & \textbf{Realizable} & \textbf{Black-Box} & & \textbf{Scale}\\ \hline

Szegedy et al. (2013) \cite{szegedy2013intriguing}, & \multirow{2}{*}{\greencheck} & \multirow{2}{*}{\redx} & \multirow{2}{*}{\redx} & \multirow{2}{*}{\redx} & \multirow{2}{*}{\redx} & \multirow{2}{*}{\redx} & \multirow{2}{*}{\greencheck} \\
Goodfellow et al. (2014) \cite{goodfellow2014explaining} & & & & & & & \\ \hline

Natural mimics, e.g.  & \multirow{2}{*}{\greencheck} & \multirow{2}{*}{\greencheck} & \multirow{2}{*}{\redx} & \multirow{2}{*}{\greencheck} & \multirow{2}{*}{\greencheck} & \multirow{2}{*}{\redx} & \multirow{2}{*}{N/A} \\
Peacock, Ringlet Butterfly & & & & & & & \\ \hline

Hayes et al. (2018)\cite{hayes2018learning} & \greencheck & \greencheck & \redx & \redx & \greencheck & \redx & \greencheck \\ \hline

Mopuri et al. (2018)a\cite{mopuri2018nag} & \greencheck & \greencheck & \redx & \redx & \greencheck & \redx & \greencheck \\ \hline

Mopuri et al. (2018)b\cite{mopuri2018ask} & \greencheck & \greencheck & \redx & \redx & \greencheck & \redx & \greencheck \\ \hline

Poursaeed et al. (2018) \cite{poursaeed2018generative} & \greencheck & \greencheck & \redx & \redx & \greencheck & \redx & \greencheck \\ \hline

Xiao et al. (2018) \cite{xiao2018generating} & \greencheck & \redx & \redx & \redx & \greencheck & \redx & \greencheck \\ \hline

Hashemi et al. (2020) \cite{hashemi2020transferable} & \greencheck & \greencheck & \redx & \redx & \greencheck & \redx & \greencheck \\ \hline

Wong et al. (2020) \cite{wong2020learning} & \greencheck & \redx & \redx & \redx & \redx & \redx & \redx \\ \hline

Liu et al. (2018) \cite{liu2018beyond} & \greencheck & \redx & \greencheck & \redx & \redx & \redx & \redx \\ \hline

Samangouei et al. (2018) \cite{samangouei2018explaingan} & \greencheck & \redx & \redx & \redx & \redx & \redx & \redx \\ \hline

Song et al. (2018) \cite{song2018constructing} & \greencheck & \redx & \greencheck & \redx & \greencheck & \redx & \redx \\ \hline

Joshi et al. (2018) \cite{joshi2018xgems} & \greencheck & \redx & \redx & \redx & \redx & \redx & \redx \\ \hline

Joshi et al. (2019) \cite{joshi2019semantic} & \greencheck & \redx & \greencheck & \redx & \redx & \redx & \redx \\ \hline

Singla et al. (2019) \cite{singla2019explanation} & \greencheck & \redx & \redx & \redx & \greencheck & \redx & \redx \\ \hline

Hu et al. (2021) \cite{hu2021naturalistic} & \greencheck & \greencheck & \greencheck & \greencheck & \redx & \redx & \redx \\ \hline

Wang et al. (2020) \cite{wang2020generating} & \greencheck & \greencheck & \greencheck & \redx & \redx & \redx & \redx \\ \hline

Kurakin et al. (2016) \cite{kurakin2016adversarial} & \greencheck & \redx & \redx & \greencheck & \greencheck & \redx & \greencheck \\ \hline

Sharif et al. (2016) \cite{sharif2016accessorize} & \greencheck & \redx & \redx & \greencheck & \greencheck & \redx & \greencheck \\ \hline

Brown et al. (2017) \cite{brown2017adversarial} & \greencheck & \greencheck & \redx & \greencheck & \greencheck & \redx & \greencheck \\ \hline

Eykholt et al. (2018) \cite{eykholt2018robust} & \greencheck & \redx & \greencheck & \greencheck & \redx & \redx & \redx \\ \hline

Athalye et al. (2018) \cite{athalye2018synthesizing} & \greencheck & \redx & \redx & \greencheck & \redx & \redx & \greencheck \\ \hline

Liu et al. (2019) \cite{liu2019perceptual} & \greencheck & \redx & \redx & \greencheck & \greencheck & \redx & \greencheck \\ \hline

Thys et al. (2019) \cite{thys2019fooling} & \greencheck & \greencheck & \redx & \greencheck & \redx & \redx & \redx \\ \hline

Kong et al. (2020) \cite{kong2020physgan} & \greencheck & \redx & \redx & \greencheck & \redx & \redx & \redx \\ \hline

Komkov et al. (2021) \cite{komkov2021advhat} & \greencheck & \greencheck & \redx & \greencheck & \redx & \redx & \redx \\ \hline

Dong et al. (2017) \cite{dong2017towards} & \greencheck & \redx & \redx & \redx & \greencheck & \redx & \greencheck \\ \hline

Geirhos et al. (2018) \cite{geirhos2018imagenet} & \redx & \redx & \redx & \redx & \redx & \redx & \greencheck \\ \hline

Leclerc et al. (2021) \cite{leclerc20213db} & \redx & \redx & \greencheck & \redx & \redx & \greencheck & \greencheck \\ \hline

Wiles et al. (2022) \cite{wiles2018discovering} & \redx & \redx & \greencheck & \redx & \greencheck & \redx & \greencheck \\ \hline

Carter et al. (2019) \cite{carter2019activation} & \greencheck & \redx & \greencheck & \redx & \redx & \greencheck & \greencheck \\ \hline

Mu et al. (2020) \cite{mu2020compositional} & \redx & \redx & \greencheck & \redx & \redx & \greencheck & \greencheck \\ \hline

Hernandez et al. (2022) \cite{hernandez2022natural} & \redx & \redx & \greencheck & \redx & \redx & \greencheck & \greencheck \\ \Xhline{3\arrayrulewidth}

\textbf{Ours} & \greencheck & \greencheck & \greencheck & \greencheck & \greencheck & \greencheck & \greencheck \\ \hline

\end{tabular}
\caption{Our feature-level attacks are uniquely versatile. Each row represents a related work (in the order in which they are presented in Section \ref{sec:related_work}.) Each column indicates a demonstrated capability of a method. Note that two methods each having a \greencheck for a capability does not imply they do so equally well. \emph{Targeted}=working for an arbitrary target class. \emph{Universal}=working for any source example. \emph{Disguised}=Perceptible and resembling something other than the target class. \emph{Physically-realizable}=working in the physical world. \emph{Transferable/black-box}=transferring to other classifiers. \emph{Copy/Paste}=useful for designing attacks in which a natural feature is pasted into a natural image.}
\label{tbl:comparisons} 
\end{table*}

\textbf{Inspiration from Nature:} Mimicry is common in nature, and sometimes, rather than holistically imitating another species, a mimic will only display particular features.
For example, many animals use adversarial eyespots to confuse predators \cite{stevens2014animal} (see Appendix \ref{app:adversarial_features_in_nature} Fig. \ref{fig:nature}a). 
Another example is the mimic octopus which imitates the patterning, but not the shape, of a banded sea snake. 
We show in Figure \ref{fig:nature}b that a ResNet50 classifies an image of one as a sea snake. 

\textbf{Generative Modeling:} 
An approach related to ours has been to train a generator or autoencoder to produce small adversarial perturbations that are applied to natural inputs. 
This has been done to synthesize imperceptible attacks that are transferable, universal, or efficient to produce \cite{hayes2018learning, mopuri2018nag, mopuri2018ask, poursaeed2018generative, xiao2018generating, hashemi2020transferable, wong2020learning}.
Rather than training a generator, ours and other works have perturbed the latents of pretrained generative models to produce perceptible alterations.
\cite{liu2018beyond} did this with a differentiable image renderer.
Others \cite{samangouei2018explaingan, song2018constructing, joshi2018xgems, joshi2019semantic, singla2019explanation, hu2021naturalistic} have used deep generative networks, and \cite{wang2020generating} aimed to create more semantically-understandable attacks by using an autoencoder with a ``disentangled'' embedding space.
Our work is different in four ways. 
(1) These works focus on small classifiers trained on simple datasets (MNIST \cite{lecun2010mnist}, Fashion MNIST \cite{xiao2017fashion}, SVHN \cite{netzer2011reading}, CelebA \cite{liu2015faceattributes}, BDD \cite{yu2018bdd100k}, INRIA \cite{dalal2005histograms}, and MPII \cite{andriluka20142d}) while we work at the ImageNet \cite{russakovsky2015imagenet} scale.  
(2) We do not simply rely on using small features or restricting the distance to a benign image to enforce the adversarialness of attacks. 
We introduce techniques that regularize the adversarial feature to be perceptible yet disguised to resemble something other than the target class. 
(3) We evaluate three distinct ways of inserting adversarial features into images.
(4) Our work is interpretability-oriented. 

\textbf{Attacks in the Physical World:} Physical-realizability demonstrates robustness. 
We show that our attacks work when printed and photographed.
This directly relates to \cite{kurakin2016adversarial} who found that pixel-space adversaries could do this to a limited extent in controlled settings.
More recently, \cite{sharif2016accessorize, brown2017adversarial, eykholt2018robust, athalye2018synthesizing, liu2019perceptual, thys2019fooling, kong2020physgan, komkov2021advhat} created adversarial clothing, stickers, or objects.
In contrast with these, we also produce attacks in the physical world that are disguised as a non-target class. 

\textbf{Adversaries and Interpretability:} 
Using adversarial examples to better interpret networks has been proposed by \cite{dong2017towards} and \cite{ tomsett2018failure}. 
We use ours to discover human-describable feature/class associations learned by a network. 
This relates to \cite{geirhos2018imagenet, leclerc20213db, wiles2018discovering} who debug networks by searching over transformations, textural changes, and feature feature alterations.
More similar to our work are \cite{carter2019activation, mu2020compositional, hernandez2022natural}, which use feature visualization \cite{olah2017feature} and network dissection \cite{bau2017network} to interpret the network. 
Each uses their interpretations to design ``copy/paste'' attacks in which one natural image pasted inside another causes an unrelated misclassification.
We add to this work with a new method to identify such adversarial features.
Unlike any previous approach, ours does so in a way that allows for targeted attacks that take into account an arbitrary distribution of source images.

\section{Methods}
\label{sec:methods}


\begin{figure*}[t!]
\centering
\includegraphics[width=0.95\linewidth]{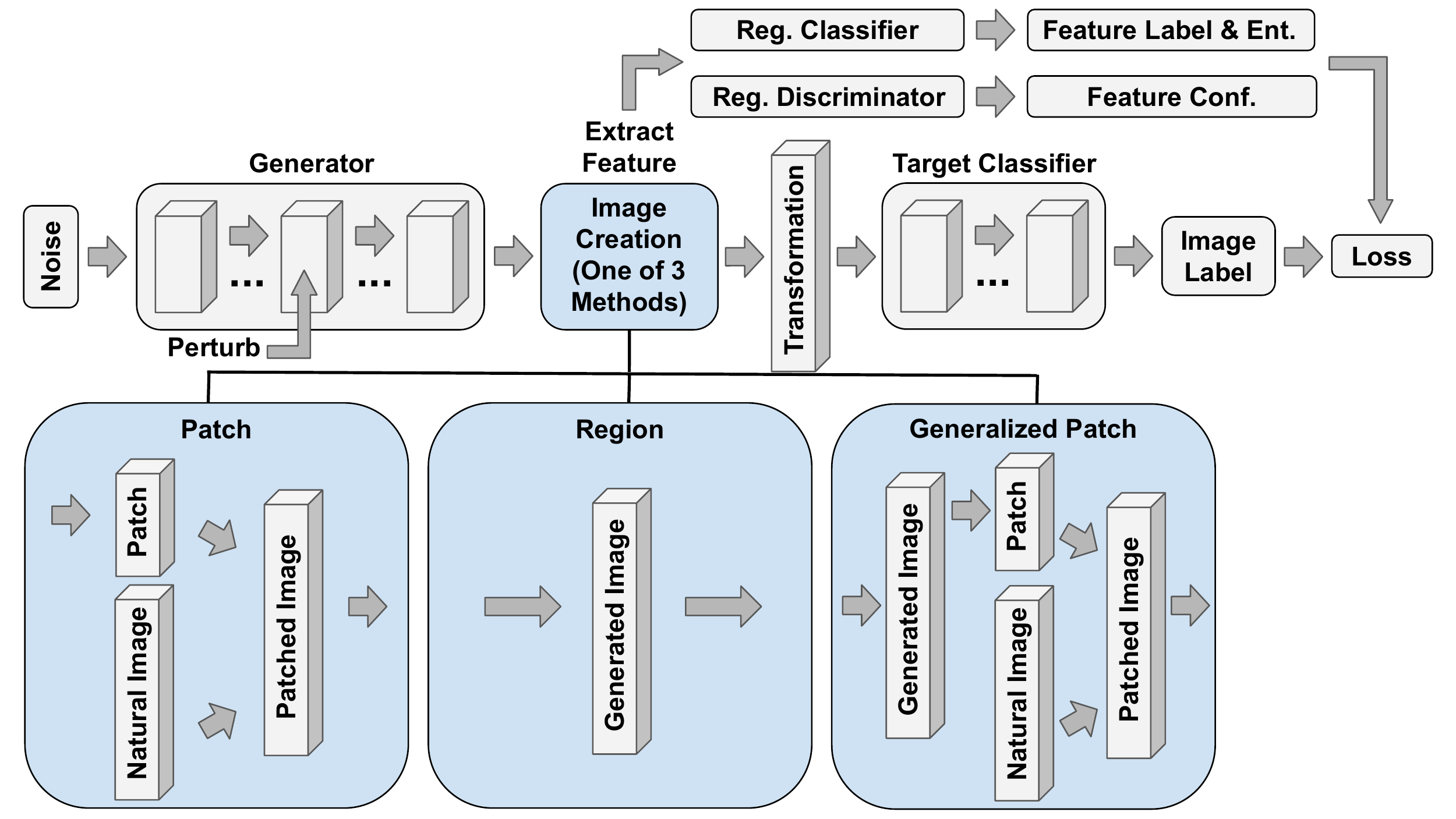}\\
\caption{Our fully-differentiable pipeline for creating feature-level attacks. In each experiment, we create either ``patch,'' ``region,'' or ``generalized patch'' attacks. The regularization terms in the loss based on an external classifier and discriminator are optional and are meant to make the inserted feature appear disguised as some non-target class.}
\label{fig:pipeline}
\end{figure*}


We adopt the ``unrestricted'' adversary paradigm \cite{song2018constructing} under which an attack is successful if the network's classification differs from an oracle's (e.g., a human). 
Our adversaries can only change a small, fixed portion of either the generator's latent or the image.
We use white-box access to the network, though we present black-box attacks based on transfer from an ensemble in Appendix \ref{app:black_box_attacks}.

Our attacks involve perturbing the latent representation in some layer of an image generator to produce an adversarial feature-level alteration.
Fig. \ref{fig:pipeline} depicts our approach.
We test three types of attacks, ``patch'', ``region'', and ``generalized patch'' (plus a fourth in Appendix \ref{app:channel_attacks} which we call ``channel'' attacks). 
We find patch attacks to generally be the most successful. 

\textbf{Patch:} We use the generator to produce a square patch that is inserted into a natural image \cite{https://doi.org/10.48550/arxiv.2206.08304}. 

\textbf{Region:} Starting with some generated image, we randomly select a square column of the latent in a generator layer which spans the channel dimension 
and replace it with a learned insertion. 
This is analogous to a square patch in the pixel representation.
We keep the insertion location fixed over training.
The modified latent is passed through the rest of the generator, producing the adversarial image.

\textbf{Generalized Patch:} These patches can be of any shape, hence the name ``generalized'' patch. 
We first generate a region attack and then extract a generalized patch from it. 
We do this by taking the absolute-valued pixel-level difference between the original and adversarial image, applying a Gaussian filter for smoothing, and creating a binary mask from the top decile of these pixel differences.
We apply this mask to the generated image to isolate the region that the perturbation altered. 
We can then treat this as a patch and overlay it onto an image in any location. 

\textbf{Basic Attacks:} For all attacks, we train a perturbation $\delta$ to the latent of the generator to minimize a loss that optimizes for both attacking the classifier and appearing interpretable:
\begin{equation} \label{eq:objective}
    \argmin_{\delta}\mathbb{E}_{x\sim \mathcal{X}, t \sim \mathcal{T}, l \sim \mathcal{L}} \;\;\;\;  L_{\textrm{x-ent}}[C(A(x,\delta,t,l)), y_\textrm{targ}] + L_{\textrm{reg}}[A(x,\delta,t,l)]
\end{equation}
with $\mathcal{X}$ a distribution over source images (e.g., a dataset or generation distribution), $\mathcal{T}$ a distribution over transformations, $\mathcal{L}$ a distribution over insertion locations (this only applies for patches and generalized patches), $C$ the target classifier, $A$ an image-generating function, $L_{\textrm{x-ent}}$ a targeted crossentropy loss for attacking the classifier, $y_{\textrm{targ}}$ the target class, and $L_{\textrm{reg}}$ a regularization loss.
The adversary has no control over $\mathcal{X}$, $\mathcal{T}$, or $\mathcal{L}$, so it must learn features that work on the network independent of any particular source image, transformation, or insertion location.
For all of our attacks, $L_{\textrm{reg}}$ contains a total variation loss, $TV(a)$, to discourage high-frequency patterns.

\begin{figure*}[t!]
\centering
\includegraphics[width=\linewidth]{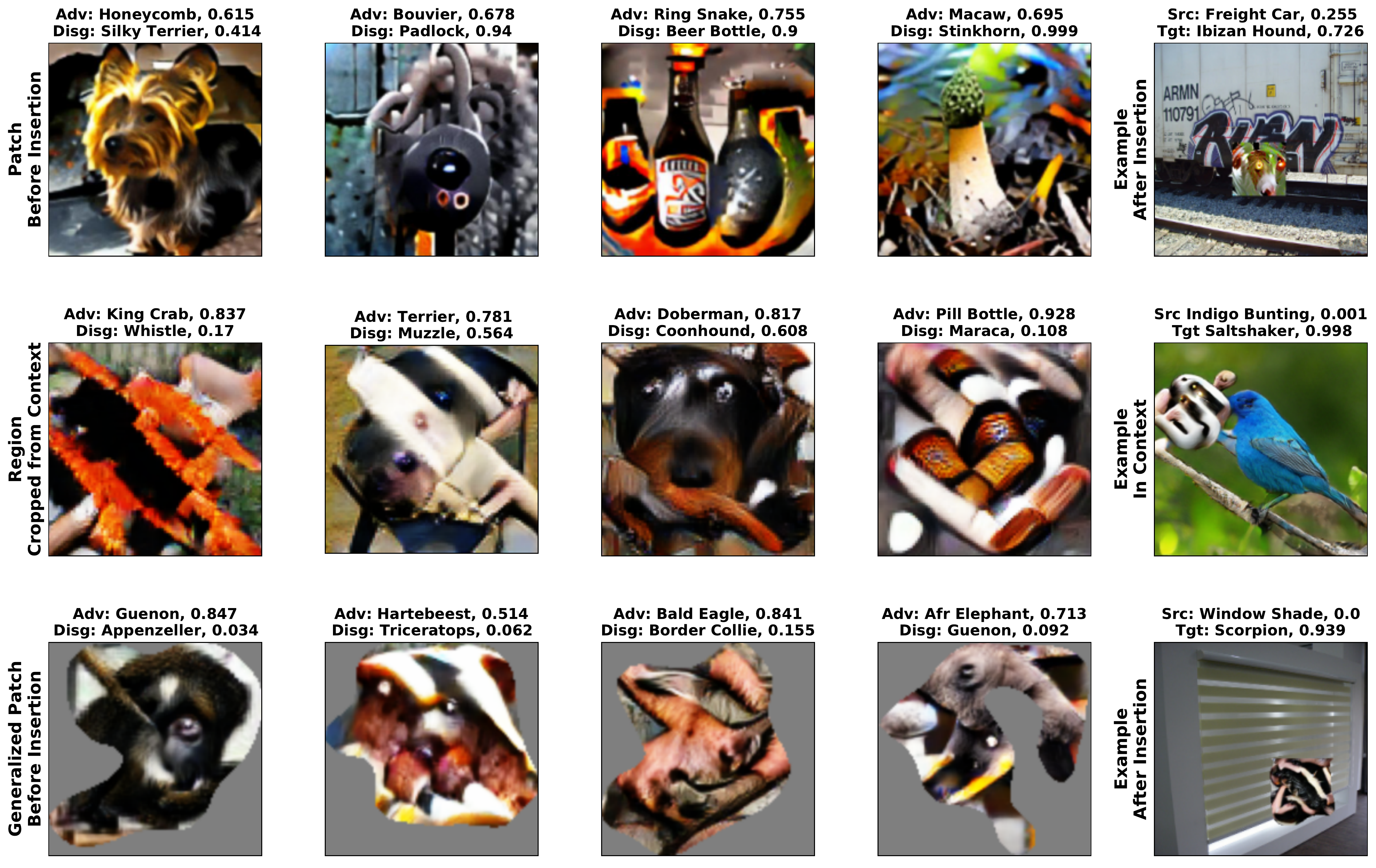}\\
\caption{Examples of targeted, universal feature-level adversaries from patch (top), region (middle), and generalized patch (bottom) attacks. The first four columns show the adversarial features. The mean target class confidence is labeled `Adv.' and is calculated under random source images (and random insertion locations for patch and generalized patch attacks). The target network's disguise class confidence for each patch or extracted generalized patch is labeled `Disg.' The final column shows examples of the features applied to images. The example image for each is labeled with its source and target class confidences.}

\label{fig:examples}
\end{figure*}

\textbf{``Disguised'' Attacks:} 
Ideally, a feature-level adversarial example should appear to a human as easily describable but should not resemble the attack's target class.
We call such attacks ``disguised.''
Here, the main goal is not to fool a human, but to help them \emph{learn} about what types of realistic features might cause the model to make a mistake.
To train these disguised attacks, we use additional terms in $L_{\textrm{reg}}$ as proxies for these two criteria.
We differentiably resize the patch or the extracted generalized patch and pass it through a GAN discriminator and auxiliary classifier.
We then add weighted terms to the regularization loss based on the discriminator's ($D$) logistic loss for classifying the input as fake, the output entropy ($H$) of some classifier ($C'$), and/or the negative of the classifier's crossentropy loss for labeling the input as the attack's target class. 
Note that $C'$ could either be the same or different than the target classifier $C$.
With all of these terms, the regularization objective is
\begin{equation} \label{eq:l_reg}
    L_{\textrm{reg}}(a) = \lambda_1 TV(a) + \underbrace{\lambda_2 L_{\textrm{logistic}}[D(P(a))] + \lambda_3 H[C'(P(a))] -\lambda_4 L_{\textrm{x-ent}}[C'(P(a), y_\textrm{targ})]}_\text{``Disguise'' Regularizers}.
\end{equation}
Here, $P(a)$ returns the extracted and resized patch from adversarial image $a$.
In order, these three new terms encourage the adversarial feature to (1) look realistic, and (2) look like some specific class, but (3) not the target class.
The choice of disguise class is left entirely to the training process.

\section{Experiments}
\label{sec:experiments}

\begin{wrapfigure}{r}{0.5\textwidth}
\centering
\includegraphics[width=\linewidth]{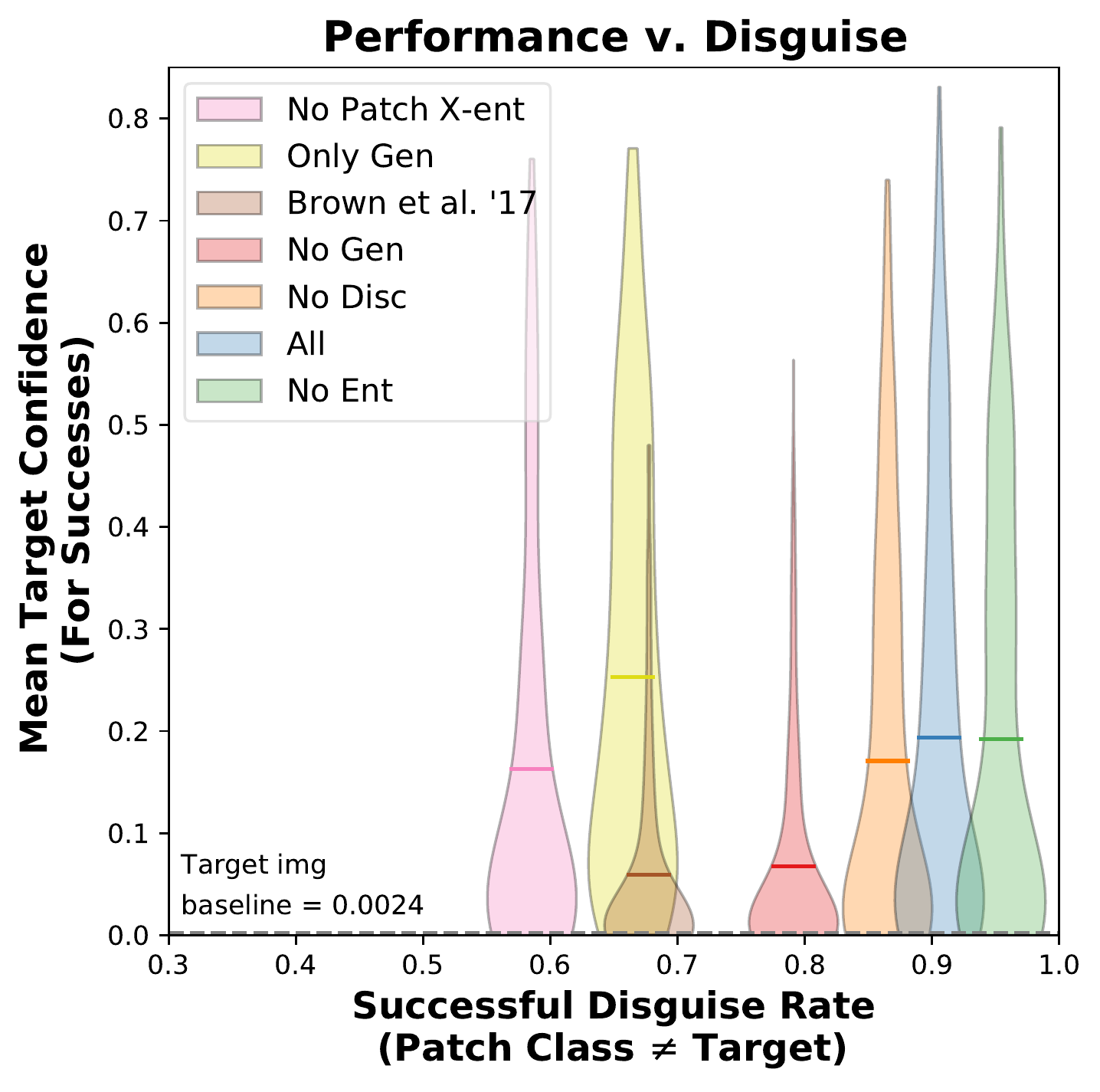}\\
\caption{Targeted, universal patch attacks compared. Successful disguise success rate (x axis) shows the proportion of attacks in which the patch was not classified by the network as the target class when viewed on its own. Mean target class confidence (y axis) gives the empirical target class confidences of 250 patch attacks. Each is an average over 100 source images. The proportion of each distribution above 0.5 gives a lower bound for the top-1 attack success rate. The mean target class confidence for using randomly sampled natural target class images as patches is 0.0024 and is shown as a thin dotted line at the bottom. }
\label{fig:violins}
\end{wrapfigure}

We use BigGAN generators from \cite{brock2018large, pretrainedbiggan}, and perturb the post-ReLU outputs of the internal `GenBlocks.' 
We also found that training slight perturbations to the BigGAN's inputs improved performance. 
We used the BigGAN discriminator and adversarially trained classifiers from \cite{robustness} for disguise regularization.
By default, we attacked a ResNet50 \cite{he2016deep}, restricting patch attacks to 1/16 of the image and region and generalized patch attacks to 1/8.
Appendix \ref{app:details} has additional details. First, in Section \ref{sec:robust_attacks} we show that these feature-level adversaries are highly robust to suggest that interpretations based on them are generalizable. 
Second, in Section \ref{sec:interpretable_copy_paste_attacks} we put these interpretations to the test and show that our feature-level adversaries can help one understand a network well enough to exploit it. 

\subsection{Robust Attacks} \label{sec:robust_attacks}

Figure \ref{fig:examples} shows examples of targeted, universal, and disguised feature-level patch (top), region (middle), and generalized patch (bottom) attacks which were each trained with all of the disguise regularization terms from Eq. \ref{eq:l_reg}.
We find the disguises to be effective, particularly for the patches (top row), but imperfect.
Appendix \ref{app:how_successful_are_our_disguises} discusses this and what it may suggest about networks and size bias.

\textbf{Performance versus Disguise:}
Here, we study our patch attacks in depth to test how effective they are at attacking the network and how successfully they can help to identify non-target-class features that can fool the network.
We compared seven different approaches. 
The first was our full approach using the generator and all disguise regularization terms from Eq. \ref{eq:l_reg}. 
The rest were ablation tests in which we omitted the generator (No Gen), the discriminator (No Disc) regularization term (No Reg), the entropy regularization term (No Ent), the crossentropy regularization term (No Patch X-ent), all three regularization terms (Only Gen), and finally the discriminator and all three regularization terms (Brown et al. '17). 
This final unregularized, pixel-level method resulted in the same approach as Brown et al. (2017) \cite{brown2017adversarial}.
For each test, all else was kept identical including penalizing total variation, training under transformations, and initializing the patch as a generator output.

For each method, we generated universal attacks with random target classes until we obtained 250 successfully ``disguised'' ones in which the resulting adversarial feature was not classified by the network as the target class when viewed on its own.
Fig. \ref{fig:violins} plots the success rate versus the distribution of target class mean confidences for each type of attack. 
For all methods, these universal attacks have variable target class confidences due in large part to the random selection of the target class.  
Some attacks are stochastically Pareto dominated by others. 
For example, the pixel-space Brown et al. (2017) attacks were the least effective at attacking the target network and had the third least disguise rate.
In other cases, there is a tradeoff between attack performance and disguise which can be controlled using the regularization terms from Eq. \ref{eq:l_reg}.
We also compare our attacks to two baselines using resized natural images from the target class and randomly sampled patches from the center of target class images.
These resulted in a mean target class confidences of 0.0024 and 0.0018 respectively.

Notably, Fig. \ref{fig:violins} does not capture everything that one might care about in these attacks.
It does not show any measure of how ``realistic'' the resulting patches look. 
In Appendix \ref{app:fooling_v_disguise}, Fig. \ref{fig:fooling_v_interpretability} plots the same target class confidence data from the $y$ axis in Fig. \ref{fig:violins} versus the disguise class label confidence from an Inception-v3 which we use as a proxy for how realistic a human would find the patch.
It suggests that the best attacks for producing patches that appear realistic are the ``All'' and ``No Disc'' methods. 
In Appendix \ref{app:printable_examples}, Figs. \ref{fig:all_printable} \ref{fig:only_gen_printable}, and \ref{fig:brown_printable} give examples of successful ``All'', ``Only Gen'', and ``Brown et al. '17'' attacks respectively. 
Because they were initialized from generator outputs, some of the ``Brown et al. '17'' attacks have a veneer-like resemblance to non-target class features.
Nonetheless, they contain higher-frequency patterns and less coherent objects in comparison to the two sets of feature-level attacks. 
We subjectively find the ``All'' attacks to be the best disguised.

\begin{figure*}[t!]
\centering
\includegraphics[width=\linewidth]{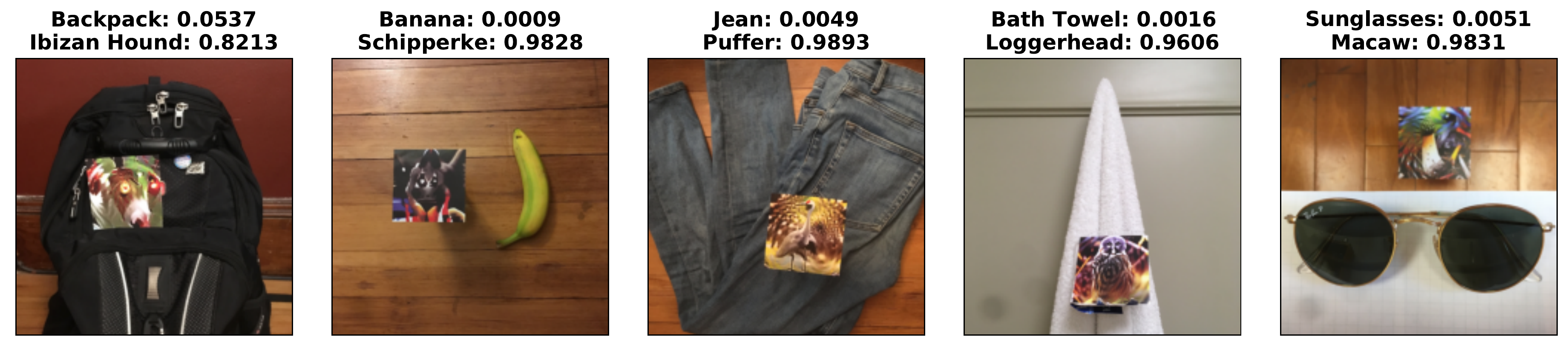}\\
\caption{Examples of targeted, disguised, universal, and physically-realizable feature-level attacks. See Appendix \ref{app:printable_examples} Fig. \ref{fig:all_printable} for full-sized versions of the patches.}
\label{fig:physical_examples}
\end{figure*}

\textbf{Physical-Realizability:}
To test their ability to transfer to the physical world, we generated 100 additional targeted, universal, and disguised adversarial patches.
We used the generator and all regularization terms (the ``All'' condition from above). 
We selected the 10 with the best mean target class confidence, printed them, and photographed each next to 9 objects from different ImageNet classes.\footnote{Backpack, banana, bath towel, lemon, jeans, spatula, sunglasses, toilet tissue, and toaster.} 
We confirmed that photographs of each object were correctly classified without a patch. 
Figure \ref{fig:physical_examples} shows successful examples.
Meanwhile, resizable and printable versions of all these patches and others are in Appendix \ref{app:printable_examples}.
The mean and standard deviation of the target class confidences for our attacks in the physical world were 0.312 and 0.318 respectively ($n=90$, not i.i.d.).
This means that these patches' mean effectiveness dropped by less than $\frac{1}{2}$ when transferring to the physical world.

\textbf{Black-Box Attacks:} In Appendix \ref{app:black_box_attacks}, we show that our targeted universal attacks can transfer from an ensemble to a held-out model.

\subsection{Interpretability} \label{sec:interpretable_copy_paste_attacks}

If an adversarial feature successfully fools the victim network, this suggests that the network associates that feature in context of a source image with the target class.
We find that our adversaries can suggest both beneficial and harmful feature-class associations. 
In Appendix \ref{app:discovering_feature_class_associations}, Fig. \ref{fig:barbershop_bikini} provides a simple example of each. 

Simply developing an interpretation, however, is easy. 
Showing that one leads to a useful understanding of the network is harder. 
One challenge in the explainable AI literature is to develop interpretations that go beyond seeming plausible and stand up to scrutiny \cite{raukur2022toward}.
Robust feature-level adversarial patches can easily be used to develop hypotheses about the network's behavior, e.g. ``The network thinks that bee features plus colorful balls implies a fly.''
But are these valid, useful interpretations of the network?
In other words, are our adversaries adversarial because of their interpretable qualities, or is it because of hidden motifs?
We verify interpretations by using our attacks to make and validate predictions about how to fool the target network with natural objects.

\textbf{Validating Interpretations with Copy/Paste Attacks:} 
A ``copy-paste'' attack is created by inserting one natural image into another to cause an unexpected misclassification. 
They are more restricted than patch attacks because the features pasted into an image must be natural objects. 
As a result, they are of high interest for physically-realizable attacks because they suggest combinations of real objects that yield unexpected classifications.
They also have precedent in the real world.
For example, subimage insertions into pornographic images have been used to evade NSFW content detectors \cite{yuan2019stealthy}.

\begin{figure*}[t!]
\centering
\begin{tabular}{cc}
\includegraphics[width=\linewidth]{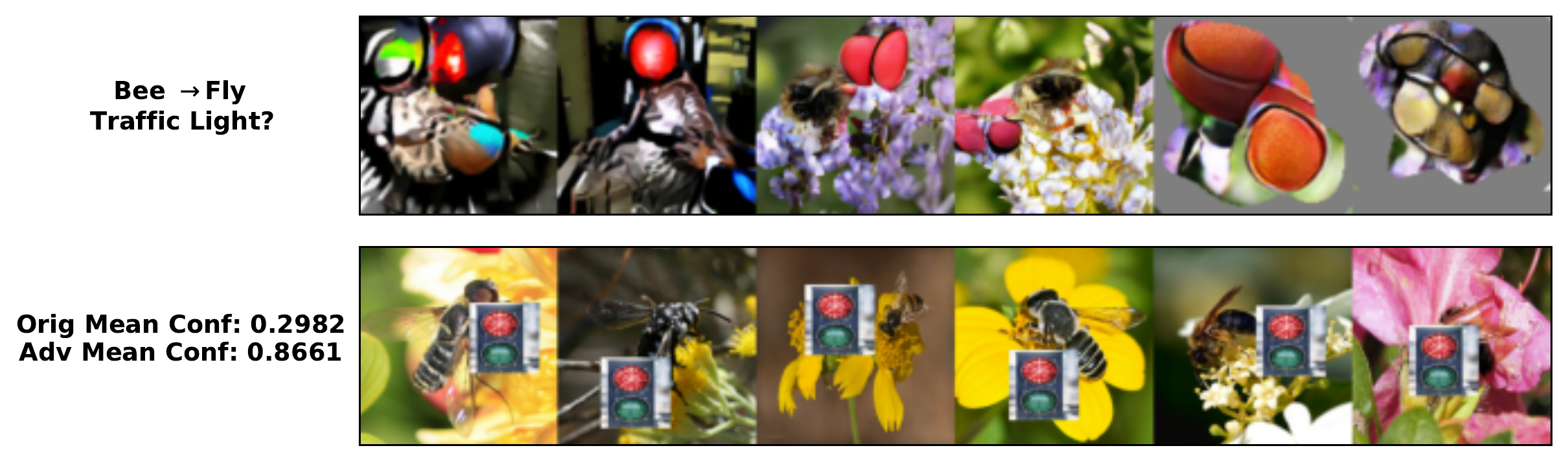}\\
\hline
\includegraphics[width=\linewidth]{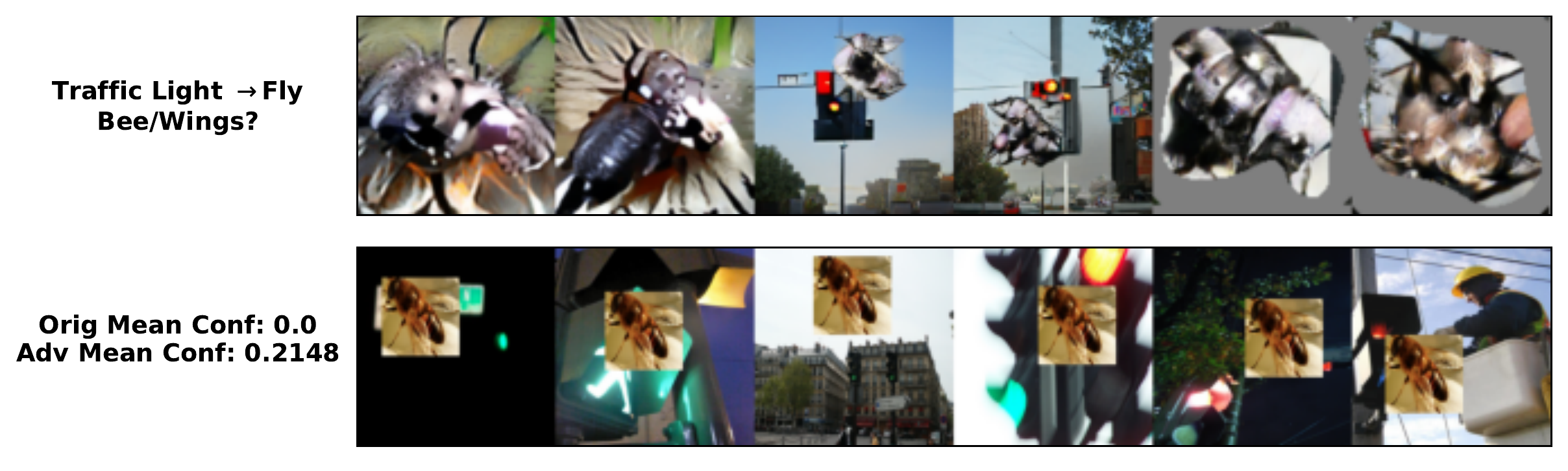}\\
\hline
\includegraphics[width=\linewidth]{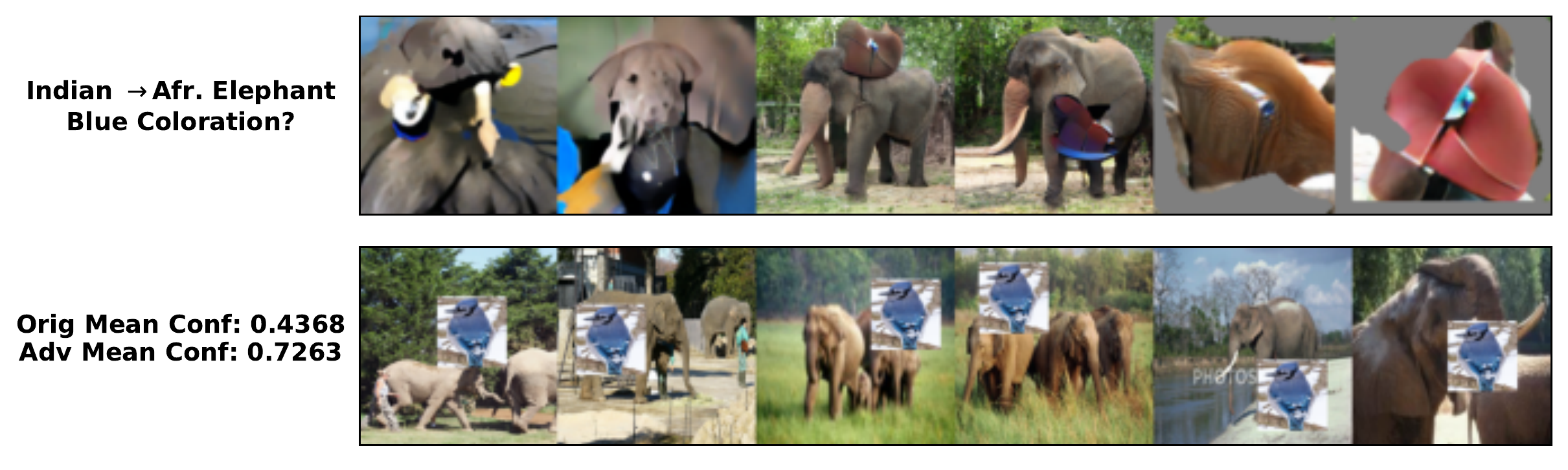}\\
\hline
\includegraphics[width=\linewidth]{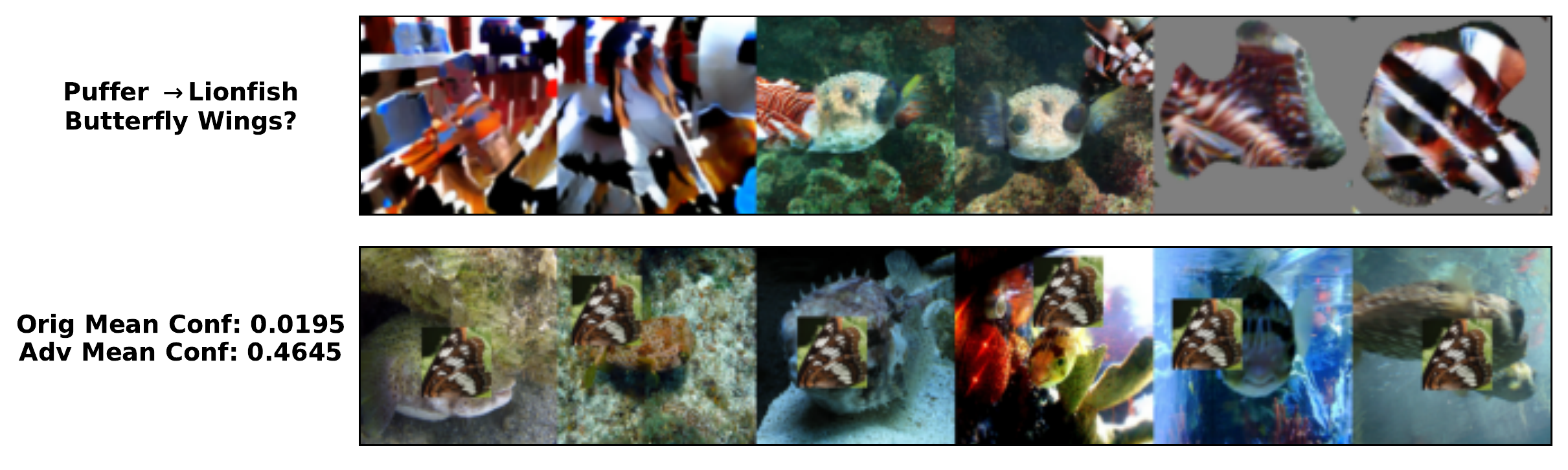}
\end{tabular}
\caption{Feature-level adversaries can guide the design of class-universal copy/paste adversarial attacks. Patch adversary pairs are on the left, region in the middle, and generalized patch on the right of each odd row. Attack examples are on each even row. We do not claim that the traffic light $\land$ bee $\to$ fly examples on row 4 are necessarily adversarial, but they demonstrate alongside the bee $\land$ traffic light $\to$ fly adversaries that the adversarial features are sensitive to the source images. For each attack except traffic light $\to$ fly, we limited ourselves to attempting 10 natural patches.}
\label{fig:copy_paste}
\end{figure*}

To develop copy/paste attacks, we select a source and target class, generate class-universal adversarial features, and manually analyze them for motifs that resemble natural objects. 
Here, we used basic attacks without the disguise regularization terms from Eq. \ref{eq:l_reg}.
We then paste images of these objects into natural images and pass them through the classifier.

Fig. \ref{fig:copy_paste} shows four types of copy/paste attacks. 
In each odd row, we show six patch, region, and generalized patch adversaries that were used to guide the design of a copy/paste attack.
In each even row are the copy/paste adversaries for the 6 (of 50) images for the source class for which the insertion resulted in the highest target class confidence increase along with the mean target class confidences before and after patch insertion for those 6. 
The success of these attacks shows their usefulness for interpreting the target network because they require that a human understands the mistake the model is making like ``Bee $\land$ Traffic Light $\to$ Fly'' well enough to manually exploit it. 
Given the differences in the adversarial features that are produced in the Bee $\to$ Fly and Traffic Light $\to$ Fly attacks, Fig. \ref{fig:copy_paste} also demonstrates how our attacks take the distribution of source images into account.

\textbf{Comparisons to Other Methods:} Three prior works \cite{carter2019activation, mu2020compositional, hernandez2022natural} have developed copy/paste attacks, also via interpretability tools.
Unlike \cite{mu2020compositional, hernandez2022natural}, our approach allows for targeted attacks. 
And unlike all three, rather than simply identifying features associated with a class, our adversaries generate adversarial features for a target class \emph{conditional} on any distribution over source images (i.e. the source class) with which the adversaries are trained.
Little work has been done on copy/paste adversaries, and thus far, methods have either not allowed for targeted attacks or have required a human in the loop.
This makes objective comparisons difficult. 
However, we provide examples of a feature-visualization based method inspired by \cite{carter2019activation} in Appendix \ref{app:copy_paste_attacks_with_class_impressions} to compare with ours. 
For the Indian $\to$ African Elephant attack, the source and target class share many features, and we find no evidence that feature visualization is able to suggest useful features for copy/paste attacks.
This suggests that our attacks' ability to take the source image distribution into account may be more helpful for discovering certain weaknesses compared to the baseline inspired by \cite{carter2019activation}.

\section{Discussion and Broader Impact} \label{sec:discussion}

\textbf{Contributions:} Here we use feature-level adversarial examples to attack and interpret deep networks in order to contribute to a more practical understanding of network vulnerabilities. 
As an attack method, our approach is versatile.
It can produce targeted, universal, disguised, physically-realizable, black-box, and copy/paste attacks at the ImageNet scale.
This method can be also used as an interpretability tool to help diagnose flaws in models. 
We ground the notion of interpretability in the ability to make predictions about combinations of natural features that will make a model fail. 
And finally, we demonstrate this through the design of targeted copy/paste attacks for any distribution over source inputs.

\textbf{Implications:} Like any work on adversarial attacks, our approach could be used maliciously to make a system fail, but we emphasize their diagnostic value. 
Understanding threats is a prerequisite to avoiding them. 
Given the robustness and versatility of our attacks, we argue that they may be valuable for continued work to address threats that systems may face in practical applications. 
There are at least two ways in which these methods can be useful.

\textbf{Adversarial Training:} The first is for adversarial training. 
Training networks on adversarial images has been shown to improve their robustness to the attacks that are used \cite{robustness}.
But this does not guarantee robustness to other types of adversarial inputs (e.g. \cite{hendrycks2021natural}).
Our feature-level attacks are categorically different from conventional pixel-level ones, and our copy/paste attacks show how networks can be fooled by novel combinations of natural objects, failures that are outside the conventional paradigm for adversarial robustness (e.g., \cite{robustness}). 
Consequently, we expect that adversarial training on broader classes of attacks such as the one we propose here will be valuable for designing more robust models.
As a promising sign, we show in Appendix \ref{app:defense_via_adversarial_training} that adversarial training is helpful against our attacks. 

\textbf{Diagnostics:} The second is for rigorously diagnosing flaws.
We show that feature-level adversaries aid the discovery of exploitable spurious feature/class associations (Fig. \ref{fig:copy_paste}) and a socially-harmful bias (Appendix \ref{app:discovering_feature_class_associations} Fig. \ref{fig:barbershop_bikini}). 
Our approach could also be extended beyond what we have demonstrated here. 
For example, our methods may be useful for feature visualization \cite{olah2017feature} of a network's internal neurons.
An analogous approach to ours can also be used in Natural Language Processing \cite{song2020universal, perez2022red}, and we are currently working on a method for this.
Furthermore, it may be valuable to use these adversaries to identify generalizable flaws in networks that humans can easily understand but with minimal human involvement.
This would be much more scalable and prevent human priors from influencing interpretations.
See \cite{casper2022diagnostics} for follow-up work involving the fully automated discovery of copy/paste attacks.

\textbf{Limitations:} A limitation of our approach is that when multiple desiderata are optimized for at the same time (e.g., universality + transformation robustness + disguise), attacks are generally less successful, more time-consuming, and require more screening to find good ones. 
This could be a bottleneck for large-scale adversarial training. 
Ultimately, this type of attack is limited by the efficiency and quality of the generator, so future work should leverage advances in generative modeling. 
Our evaluation method is also limited to a proof-of-concept for the design of copy/paste attacks.
Future work should evaluate this more rigorously. 
We are currently working toward developing a benchmark for interpretability tools based on their ability to aid a human in rediscovering trojans \cite{geigel2013neural} that have been implanted into a model. 

\textbf{Conclusion:} As AI becomes increasingly capable, it becomes more important to design models that are reliable.
Each of the 11 proposals for building safe AI outlined in \cite{hubinger2020overview} explicitly calls for adversarial robustness and/or interpretability tools, and recent work from \cite{ziegler2022adversarial} on high-stakes reliability in AI found that interpretability tools strengthened their ability to produce inputs for adversarial training. 
Given the close relationship between interpretability and adversarial robustness, continued study of the connections between them will be key for building safer AI systems.

\section*{Acknowledgments}
We thank Cassidy Laidlaw, Miles Turpin, Will Xiao, and Alexander Davies for insightful discussions and feedback and Kaivu Hariharan for help with coding. This work was conducted in part with funding from the Harvard Undergraduate Office of Research and Fellowships. 

\bibliographystyle{plain}
\bibliography{bibliography}

\begin{thebibliography}{10}

\bibitem{andriluka20142d}
Mykhaylo Andriluka, Leonid Pishchulin, Peter Gehler, and Bernt Schiele.
\newblock 2d human pose estimation: New benchmark and state of the art
  analysis.
\newblock In {\em Proceedings of the IEEE Conference on computer Vision and
  Pattern Recognition}, pages 3686--3693, 2014.

\bibitem{athalye2018synthesizing}
Anish Athalye, Logan Engstrom, Andrew Ilyas, and Kevin Kwok.
\newblock Synthesizing robust adversarial examples.
\newblock In {\em International conference on machine learning}, pages
  284--293. PMLR, 2018.

\bibitem{bau2017network}
David Bau, Bolei Zhou, Aditya Khosla, Aude Oliva, and Antonio Torralba.
\newblock Network dissection: Quantifying interpretability of deep visual
  representations.
\newblock In {\em Proceedings of the IEEE conference on computer vision and
  pattern recognition}, pages 6541--6549, 2017.

\bibitem{bhattad2019unrestricted}
Anand Bhattad, Min~Jin Chong, Kaizhao Liang, Bo~Li, and David~A Forsyth.
\newblock Unrestricted adversarial examples via semantic manipulation.
\newblock {\em arXiv preprint arXiv:1904.06347}, 2019.

\bibitem{brock2018large}
Andrew Brock, Jeff Donahue, and Karen Simonyan.
\newblock Large scale gan training for high fidelity natural image synthesis.
\newblock {\em arXiv preprint arXiv:1809.11096}, 2018.

\bibitem{brown2017adversarial}
Tom~B Brown, Dandelion Man{\'e}, Aurko Roy, Mart{\'\i}n Abadi, and Justin
  Gilmer.
\newblock Adversarial patch.
\newblock {\em arXiv preprint arXiv:1712.09665}, 2017.

\bibitem{carter2019activation}
Shan Carter, Zan Armstrong, Ludwig Schubert, Ian Johnson, and Chris Olah.
\newblock Activation atlas.
\newblock {\em Distill}, 4(3):e15, 2019.

\bibitem{casper2022diagnostics}
Stephen Casper, Kaivalya Hariharan, and Dylan Hadfield-Menell.
\newblock Diagnostics for deep neural networks with automated copy/paste
  attacks.
\newblock In {\em NeurIPS ML Safety Workshop}, 2022.

\bibitem{dalal2005histograms}
Navneet Dalal and Bill Triggs.
\newblock Histograms of oriented gradients for human detection.
\newblock In {\em 2005 IEEE computer society conference on computer vision and
  pattern recognition (CVPR'05)}, volume~1, pages 886--893. Ieee, 2005.

\bibitem{dapello2020simulating}
Joel Dapello, Tiago Marques, Martin Schrimpf, Franziska Geiger, David~D Cox,
  and James~J DiCarlo.
\newblock Simulating a primary visual cortex at the front of cnns improves
  robustness to image perturbations.
\newblock {\em BioRxiv}, 2020.

\bibitem{dong2017towards}
Yinpeng Dong, Hang Su, Jun Zhu, and Fan Bao.
\newblock Towards interpretable deep neural networks by leveraging adversarial
  examples.
\newblock {\em arXiv preprint arXiv:1708.05493}, 2017.

\bibitem{dosovitskiy2020image}
Alexey Dosovitskiy, Lucas Beyer, Alexander Kolesnikov, Dirk Weissenborn,
  Xiaohua Zhai, Thomas Unterthiner, Mostafa Dehghani, Matthias Minderer, Georg
  Heigold, Sylvain Gelly, et~al.
\newblock An image is worth 16x16 words: Transformers for image recognition at
  scale.
\newblock {\em arXiv preprint arXiv:2010.11929}, 2020.

\bibitem{robustness}
Logan Engstrom, Andrew Ilyas, Hadi Salman, Shibani Santurkar, and Dimitris
  Tsipras.
\newblock Robustness (python library), 2019.

\bibitem{engstrom2019adversarial}
Logan Engstrom, Andrew Ilyas, Shibani Santurkar, Dimitris Tsipras, Brandon
  Tran, and Aleksander Madry.
\newblock Adversarial robustness as a prior for learned representations.
\newblock {\em arXiv preprint arXiv:1906.00945}, 2019.

\bibitem{eykholt2018robust}
Kevin Eykholt, Ivan Evtimov, Earlence Fernandes, Bo~Li, Amir Rahmati, Chaowei
  Xiao, Atul Prakash, Tadayoshi Kohno, and Dawn Song.
\newblock Robust physical-world attacks on deep learning visual classification.
\newblock In {\em Proceedings of the IEEE conference on computer vision and
  pattern recognition}, pages 1625--1634, 2018.

\bibitem{geigel2013neural}
Arturo Geigel.
\newblock Neural network trojan.
\newblock {\em Journal of Computer Security}, 21(2):191--232, 2013.

\bibitem{geirhos2018imagenet}
Robert Geirhos, Patricia Rubisch, Claudio Michaelis, Matthias Bethge, Felix~A
  Wichmann, and Wieland Brendel.
\newblock Imagenet-trained cnns are biased towards texture; increasing shape
  bias improves accuracy and robustness.
\newblock {\em arXiv preprint arXiv:1811.12231}, 2018.

\bibitem{goodfellow2014explaining}
Ian~J Goodfellow, Jonathon Shlens, and Christian Szegedy.
\newblock Explaining and harnessing adversarial examples.
\newblock {\em arXiv preprint arXiv:1412.6572}, 2014.

\bibitem{hashemi2020transferable}
Atiye~Sadat Hashemi, Andreas B{\"a}r, Saeed Mozaffari, and Tim Fingscheidt.
\newblock Transferable universal adversarial perturbations using generative
  models.
\newblock {\em arXiv preprint arXiv:2010.14919}, 2020.

\bibitem{hayes2018learning}
Jamie Hayes and George Danezis.
\newblock Learning universal adversarial perturbations with generative models.
\newblock In {\em 2018 IEEE Security and Privacy Workshops (SPW)}, pages
  43--49. IEEE, 2018.

\bibitem{he2016deep}
Kaiming He, Xiangyu Zhang, Shaoqing Ren, and Jian Sun.
\newblock Deep residual learning for image recognition.
\newblock In {\em Proceedings of the IEEE conference on computer vision and
  pattern recognition}, pages 770--778, 2016.

\bibitem{hendrycks2021natural}
Dan Hendrycks, Kevin Zhao, Steven Basart, Jacob Steinhardt, and Dawn Song.
\newblock Natural adversarial examples.
\newblock In {\em Proceedings of the IEEE/CVF Conference on Computer Vision and
  Pattern Recognition}, pages 15262--15271, 2021.

\bibitem{hernandez2022natural}
Evan Hernandez, Sarah Schwettmann, David Bau, Teona Bagashvili, Antonio
  Torralba, and Jacob Andreas.
\newblock Natural language descriptions of deep visual features.
\newblock {\em arXiv preprint arXiv:2201.11114}, 2022.

\bibitem{hu2021naturalistic}
Yu-Chih-Tuan Hu, Bo-Han Kung, Daniel~Stanley Tan, Jun-Cheng Chen, Kai-Lung Hua,
  and Wen-Huang Cheng.
\newblock Naturalistic physical adversarial patch for object detectors.
\newblock In {\em Proceedings of the IEEE/CVF International Conference on
  Computer Vision}, pages 7848--7857, 2021.

\bibitem{huang2017densely}
Gao Huang, Zhuang Liu, Laurens Van Der~Maaten, and Kilian~Q Weinberger.
\newblock Densely connected convolutional networks.
\newblock In {\em Proceedings of the IEEE conference on computer vision and
  pattern recognition}, pages 4700--4708, 2017.

\bibitem{hubinger2020overview}
Evan Hubinger.
\newblock An overview of 11 proposals for building safe advanced ai.
\newblock {\em arXiv preprint arXiv:2012.07532}, 2020.

\bibitem{hullman2011visualization}
Jessica Hullman and Nick Diakopoulos.
\newblock Visualization rhetoric: Framing effects in narrative visualization.
\newblock {\em IEEE transactions on visualization and computer graphics},
  17(12):2231--2240, 2011.

\bibitem{joshi2019semantic}
Ameya Joshi, Amitangshu Mukherjee, Soumik Sarkar, and Chinmay Hegde.
\newblock Semantic adversarial attacks: Parametric transformations that fool
  deep classifiers.
\newblock In {\em Proceedings of the IEEE/CVF International Conference on
  Computer Vision}, pages 4773--4783, 2019.

\bibitem{joshi2018xgems}
Shalmali Joshi, Oluwasanmi Koyejo, Been Kim, and Joydeep Ghosh.
\newblock xgems: Generating examplars to explain black-box models.
\newblock {\em arXiv preprint arXiv:1806.08867}, 2018.

\bibitem{lucent}
Lim Kiat.
\newblock Lucent.
\newblock \url{https://github.com/greentfrapp/lucent}, 2019.

\bibitem{komkov2021advhat}
Stepan Komkov and Aleksandr Petiushko.
\newblock Advhat: Real-world adversarial attack on arcface face id system.
\newblock In {\em 2020 25th International Conference on Pattern Recognition
  (ICPR)}, pages 819--826. IEEE, 2021.

\bibitem{kong2020physgan}
Zelun Kong, Junfeng Guo, Ang Li, and Cong Liu.
\newblock Physgan: Generating physical-world-resilient adversarial examples for
  autonomous driving.
\newblock In {\em Proceedings of the IEEE/CVF Conference on Computer Vision and
  Pattern Recognition}, pages 14254--14263, 2020.

\bibitem{krizhevsky2012imagenet}
Alex Krizhevsky, Ilya Sutskever, and Geoffrey~E Hinton.
\newblock Imagenet classification with deep convolutional neural networks.
\newblock {\em Advances in neural information processing systems},
  25:1097--1105, 2012.

\bibitem{kurakin2016adversarial}
Alexey Kurakin, Ian Goodfellow, Samy Bengio, et~al.
\newblock Adversarial examples in the physical world, 2016.

\bibitem{leclerc20213db}
Guillaume Leclerc, Hadi Salman, Andrew Ilyas, Sai Vemprala, Logan Engstrom,
  Vibhav Vineet, Kai Xiao, Pengchuan Zhang, Shibani Santurkar, Greg Yang,
  et~al.
\newblock 3db: A framework for debugging computer vision models.
\newblock {\em arXiv preprint arXiv:2106.03805}, 2021.

\bibitem{lecun2010mnist}
Yann LeCun, Corinna Cortes, and CJ~Burges.
\newblock Mnist handwritten digit database.
\newblock {\em ATT Labs [Online]. Available: http://yann.lecun.com/exdb/mnist},
  2, 2010.

\bibitem{liu2019perceptual}
Aishan Liu, Xianglong Liu, Jiaxin Fan, Yuqing Ma, Anlan Zhang, Huiyuan Xie, and
  Dacheng Tao.
\newblock Perceptual-sensitive gan for generating adversarial patches.
\newblock 33(01):1028--1035, 2019.

\bibitem{liu2018beyond}
Hsueh-Ti~Derek Liu, Michael Tao, Chun-Liang Li, Derek Nowrouzezahrai, and Alec
  Jacobson.
\newblock Beyond pixel norm-balls: Parametric adversaries using an analytically
  differentiable renderer.
\newblock {\em arXiv preprint arXiv:1808.02651}, 2018.

\bibitem{liu2015faceattributes}
Ziwei Liu, Ping Luo, Xiaogang Wang, and Xiaoou Tang.
\newblock Deep learning face attributes in the wild.
\newblock In {\em Proceedings of International Conference on Computer Vision
  (ICCV)}, December 2015.

\bibitem{vit}
Luke Melas.
\newblock Pytorch-pretrained-vit.
\newblock \url{https://github.com/lukemelas/PyTorch-Pretrained-ViT}, 2020.

\bibitem{mopuri2018nag}
Konda~Reddy Mopuri, Utkarsh Ojha, Utsav Garg, and R~Venkatesh Babu.
\newblock Nag: Network for adversary generation.
\newblock In {\em Proceedings of the IEEE Conference on Computer Vision and
  Pattern Recognition}, pages 742--751, 2018.

\bibitem{mopuri2018ask}
Konda~Reddy Mopuri, Phani~Krishna Uppala, and R~Venkatesh Babu.
\newblock Ask, acquire, and attack: Data-free uap generation using class
  impressions.
\newblock In {\em Proceedings of the European Conference on Computer Vision
  (ECCV)}, pages 19--34, 2018.

\bibitem{mu2020compositional}
Jesse Mu and Jacob Andreas.
\newblock Compositional explanations of neurons.
\newblock {\em arXiv preprint arXiv:2006.14032}, 2020.

\bibitem{netzer2011reading}
Yuval Netzer, Tao Wang, Adam Coates, Alessandro Bissacco, Bo~Wu, and Andrew~Y
  Ng.
\newblock Reading digits in natural images with unsupervised feature learning.
\newblock {\em google research}, 2011.

\bibitem{norman2001dynamic}
Mark~D Norman, Julian Finn, and Tom Tregenza.
\newblock Dynamic mimicry in an indo--malayan octopus.
\newblock {\em Proceedings of the Royal Society of London. Series B: Biological
  Sciences}, 268(1478):1755--1758, 2001.

\bibitem{ntsb2018collision}
National Transportation Safety~Board NTSB.
\newblock Collision between vehicle controlled by developmental automated
  driving system and pedestrian.
\newblock {\em ntsb}, 2018.

\bibitem{olah2017feature}
Chris Olah, Alexander Mordvintsev, and Ludwig Schubert.
\newblock Feature visualization.
\newblock {\em Distill}, 2(11):e7, 2017.

\bibitem{papernot2016transferability}
Nicolas Papernot, Patrick McDaniel, and Ian Goodfellow.
\newblock Transferability in machine learning: from phenomena to black-box
  attacks using adversarial samples.
\newblock {\em arXiv preprint arXiv:1605.07277}, 2016.

\bibitem{NEURIPS2019_9015}
Adam Paszke, Sam Gross, Francisco Massa, Adam Lerer, James Bradbury, Gregory
  Chanan, Trevor Killeen, Zeming Lin, Natalia Gimelshein, Luca Antiga, Alban
  Desmaison, Andreas Kopf, Edward Yang, Zachary DeVito, Martin Raison, Alykhan
  Tejani, Sasank Chilamkurthy, Benoit Steiner, Lu~Fang, Junjie Bai, and Soumith
  Chintala.
\newblock Pytorch: An imperative style, high-performance deep learning library.
\newblock In H.~Wallach, H.~Larochelle, A.~Beygelzimer, F.~d\textquotesingle
  Alch\'{e}-Buc, E.~Fox, and R.~Garnett, editors, {\em Advances in Neural
  Information Processing Systems 32}, pages 8024--8035. Curran Associates,
  Inc., 2019.

\bibitem{perez2022red}
Ethan Perez, Saffron Huang, Francis Song, Trevor Cai, Roman Ring, John
  Aslanides, Amelia Glaese, Nat McAleese, and Geoffrey Irving.
\newblock Red teaming language models with language models.
\newblock {\em arXiv preprint arXiv:2202.03286}, 2022.

\bibitem{poursaeed2018generative}
Omid Poursaeed, Isay Katsman, Bicheng Gao, and Serge Belongie.
\newblock Generative adversarial perturbations.
\newblock In {\em Proceedings of the IEEE Conference on Computer Vision and
  Pattern Recognition}, pages 4422--4431, 2018.

\bibitem{raukur2022toward}
Tilman R{\"a}ukur, Anson Ho, Stephen Casper, and Dylan Hadfield-Menell.
\newblock Toward transparent ai: A survey on interpreting the inner structures
  of deep neural networks.
\newblock {\em arXiv preprint arXiv:2207.13243}, 2022.

\bibitem{russakovsky2015imagenet}
Olga Russakovsky, Jia Deng, Hao Su, Jonathan Krause, Sanjeev Satheesh, Sean Ma,
  Zhiheng Huang, Andrej Karpathy, Aditya Khosla, Michael Bernstein, et~al.
\newblock Imagenet large scale visual recognition challenge.
\newblock {\em International journal of computer vision}, 115(3):211--252,
  2015.

\bibitem{salman2020adversarially}
Hadi Salman, Andrew Ilyas, Logan Engstrom, Ashish Kapoor, and Aleksander Madry.
\newblock Do adversarially robust imagenet models transfer better?
\newblock In {\em ArXiv preprint arXiv:2007.08489}, 2020.

\bibitem{samangouei2018explaingan}
Pouya Samangouei, Ardavan Saeedi, Liam Nakagawa, and Nathan Silberman.
\newblock Explaingan: Model explanation via decision boundary crossing
  transformations.
\newblock In {\em Proceedings of the European Conference on Computer Vision
  (ECCV)}, pages 666--681, 2018.

\bibitem{schrimpf2020integrative}
Martin Schrimpf, Jonas Kubilius, Michael~J Lee, N~Apurva~Ratan Murty, Robert
  Ajemian, and James~J DiCarlo.
\newblock Integrative benchmarking to advance neurally mechanistic models of
  human intelligence.
\newblock {\em Neuron}, 108(3):413--423, 2020.

\bibitem{sharif2016accessorize}
Mahmood Sharif, Sruti Bhagavatula, Lujo Bauer, and Michael~K Reiter.
\newblock Accessorize to a crime: Real and stealthy attacks on state-of-the-art
  face recognition.
\newblock In {\em Proceedings of the 2016 acm sigsac conference on computer and
  communications security}, pages 1528--1540, 2016.

\bibitem{https://doi.org/10.48550/arxiv.2206.08304}
Abhijith Sharma, Yijun Bian, Phil Munz, and Apurva Narayan.
\newblock Adversarial patch attacks and defences in vision-based tasks: A
  survey, 2022.

\bibitem{simonyan2014very}
Karen Simonyan and Andrew Zisserman.
\newblock Very deep convolutional networks for large-scale image recognition.
\newblock {\em arXiv preprint arXiv:1409.1556}, 2014.

\bibitem{singla2019explanation}
Sumedha Singla, Brian Pollack, Junxiang Chen, and Kayhan Batmanghelich.
\newblock Explanation by progressive exaggeration.
\newblock {\em arXiv preprint arXiv:1911.00483}, 2019.

\bibitem{lpipstorch}
Uchida So and Ihor Durnopianov.
\newblock Lpips pytorch.
\newblock \url{https://github.com/S-aiueo32/lpips-pytorch}, 2019.

\bibitem{song2020universal}
Liwei Song, Xinwei Yu, Hsuan-Tung Peng, and Karthik Narasimhan.
\newblock Universal adversarial attacks with natural triggers for text
  classification.
\newblock {\em arXiv preprint arXiv:2005.00174}, 2020.

\bibitem{song2018constructing}
Yang Song, Rui Shu, Nate Kushman, and Stefano Ermon.
\newblock Constructing unrestricted adversarial examples with generative
  models.
\newblock {\em arXiv preprint arXiv:1805.07894}, 2018.

\bibitem{stevens2014animal}
Martin Stevens and Graeme~D Ruxton.
\newblock Do animal eyespots really mimic eyes?
\newblock {\em Current Zoology}, 60(1), 2014.

\bibitem{szegedy2016rethinking}
Christian Szegedy, Vincent Vanhoucke, Sergey Ioffe, Jon Shlens, and Zbigniew
  Wojna.
\newblock Rethinking the inception architecture for computer vision.
\newblock In {\em Proceedings of the IEEE conference on computer vision and
  pattern recognition}, pages 2818--2826, 2016.

\bibitem{szegedy2013intriguing}
Christian Szegedy, Wojciech Zaremba, Ilya Sutskever, Joan Bruna, Dumitru Erhan,
  Ian Goodfellow, and Rob Fergus.
\newblock Intriguing properties of neural networks.
\newblock {\em arXiv preprint arXiv:1312.6199}, 2013.

\bibitem{thys2019fooling}
Simen Thys, Wiebe Van~Ranst, and Toon Goedem{\'e}.
\newblock Fooling automated surveillance cameras: adversarial patches to attack
  person detection.
\newblock In {\em Proceedings of the IEEE/CVF conference on computer vision and
  pattern recognition workshops}, pages 0--0, 2019.

\bibitem{tomsett2018failure}
Richard Tomsett, Amy Widdicombe, Tianwei Xing, Supriyo Chakraborty, Simon
  Julier, Prudhvi Gurram, Raghuveer Rao, and Mani Srivastava.
\newblock Why the failure? how adversarial examples can provide insights for
  interpretable machine learning.
\newblock In {\em 2018 21st International Conference on Information Fusion
  (FUSION)}, pages 838--845. IEEE, 2018.

\bibitem{wang2020generating}
Shuo Wang, Shangyu Chen, Tianle Chen, Surya Nepal, Carsten Rudolph, and Marthie
  Grobler.
\newblock Generating semantic adversarial examples via feature manipulation.
\newblock {\em arXiv preprint arXiv:2001.02297}, 2020.

\bibitem{wiles2018discovering}
Olivia Wiles, Isabela Albuquerque, and Sven Gowal.
\newblock Discovering bugs in vision models using off-the-shelf image
  generation and captioning, 2022.

\bibitem{pretrainedbiggan}
Thomas Wolf.
\newblock Pytorch pretrained biggan.
\newblock \url{https://github.com/huggingface/pytorch-pretrained-BigGAN}, 2018.

\bibitem{wong2020learning}
Eric Wong and J~Zico Kolter.
\newblock Learning perturbation sets for robust machine learning.
\newblock {\em arXiv preprint arXiv:2007.08450}, 2020.

\bibitem{xiao2018generating}
Chaowei Xiao, Bo~Li, Jun-Yan Zhu, Warren He, Mingyan Liu, and Dawn Song.
\newblock Generating adversarial examples with adversarial networks.
\newblock {\em arXiv preprint arXiv:1801.02610}, 2018.

\bibitem{xiao2017fashion}
Han Xiao, Kashif Rasul, and Roland Vollgraf.
\newblock Fashion-mnist: a novel image dataset for benchmarking machine
  learning algorithms.
\newblock {\em arXiv preprint arXiv:1708.07747}, 2017.

\bibitem{yu2018bdd100k}
Fisher Yu, Wenqi Xian, Yingying Chen, Fangchen Liu, Mike Liao, Vashisht
  Madhavan, and Trevor Darrell.
\newblock Bdd100k: A diverse driving video database with scalable annotation
  tooling.
\newblock {\em arXiv preprint arXiv:1805.04687}, 2(5):6, 2018.

\bibitem{yuan2019stealthy}
Kan Yuan, Di~Tang, Xiaojing Liao, XiaoFeng Wang, Xuan Feng, Yi~Chen, Menghan
  Sun, Haoran Lu, and Kehuan Zhang.
\newblock Stealthy porn: Understanding real-world adversarial images for
  illicit online promotion.
\newblock In {\em 2019 IEEE Symposium on Security and Privacy (SP)}, pages
  952--966. IEEE, 2019.

\bibitem{zhang2018unreasonable}
Richard Zhang, Phillip Isola, Alexei~A Efros, Eli Shechtman, and Oliver Wang.
\newblock The unreasonable effectiveness of deep features as a perceptual
  metric.
\newblock In {\em Proceedings of the IEEE conference on computer vision and
  pattern recognition}, pages 586--595, 2018.

\bibitem{ziegler2022adversarial}
Daniel~M Ziegler, Seraphina Nix, Lawrence Chan, Tim Bauman, Peter
  Schmidt-Nielsen, Tao Lin, Adam Scherlis, Noa Nabeshima, Ben Weinstein-Raun,
  Daniel de~Haas, et~al.
\newblock Adversarial training for high-stakes reliability.
\newblock {\em arXiv preprint arXiv:2205.01663}, 2022.

\end{thebibliography}

\newpage
\appendix
\section{Appendix}

\subsection{Adversarial Features in Nature}
\label{app:adversarial_features_in_nature}
\begin{figure*}[t!]
\centering
\begin{tabular}{c|c}
    \includegraphics[width=.25\linewidth]{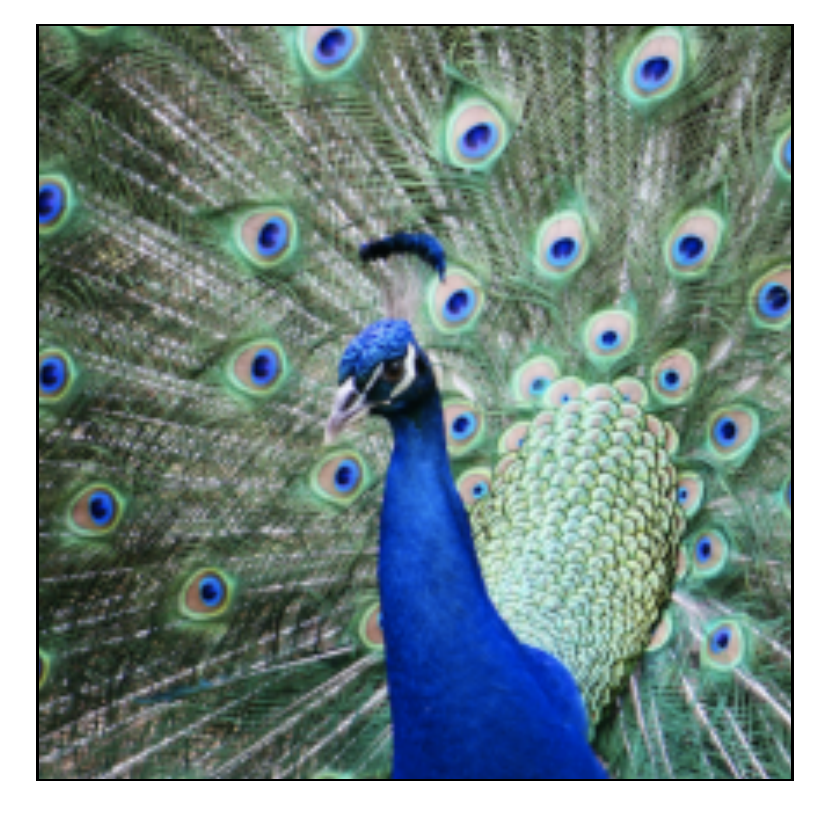}
    \includegraphics[width=.25\linewidth]{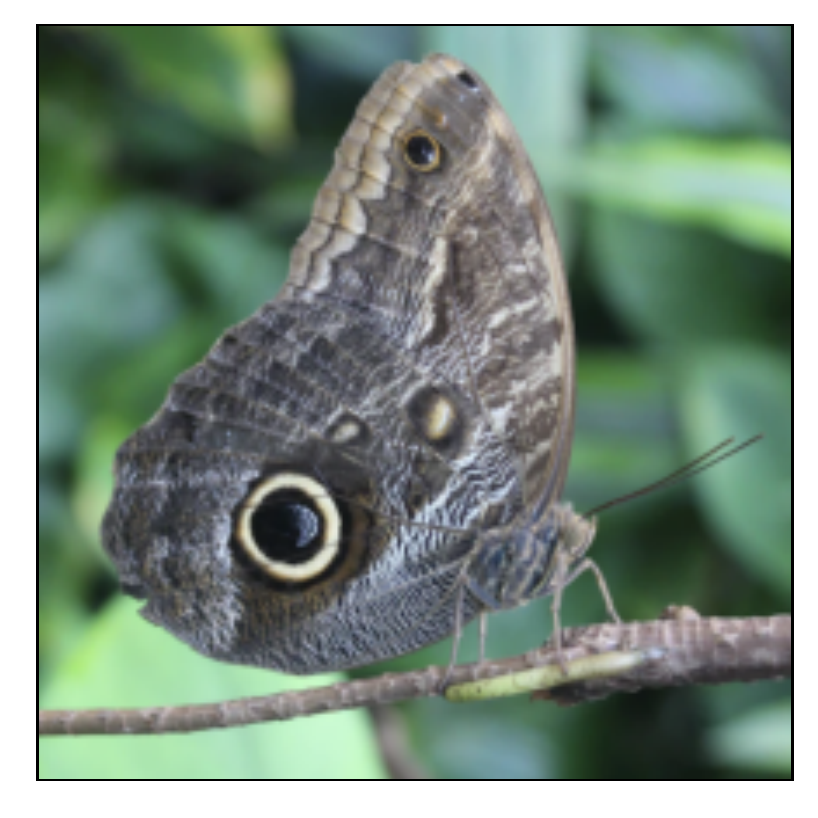} &
	\includegraphics[width=.4125\linewidth]{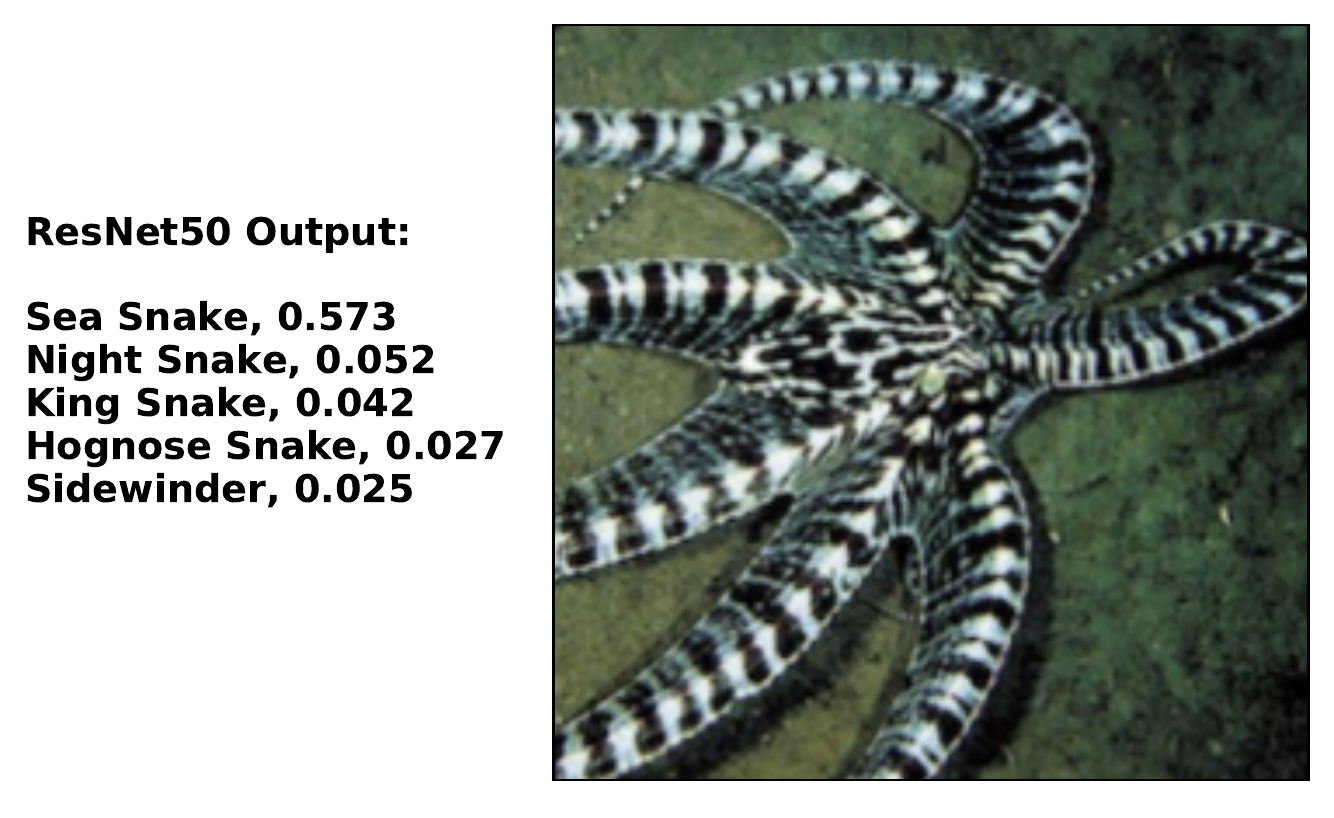}\\
	(a) & (b) \\
\end{tabular}
\caption{Examples of robust feature-level adversaries in nature. (a) A peacock and butterfly with adversarial ``eyespots.'' (b) A `mimic octopus' (image from \cite{norman2001dynamic}) mimics a banded sea snake's patterning and is classified as one by a ResNet50.}
\label{fig:nature}
\end{figure*}

Fig. \ref{fig:nature} shows examples of robust feature-level adversaries in nature.

\subsection{Methodological Details}
\label{app:details}

We found that performing crossentropy and entropy regularization (Eq. \ref{eq:l_reg}) for disguise using adversarially-trained auxiliary classifiers produced more easily interpretable results.
This aligns with findings that adversarially-trained networks tend to learn more interpretable representations \cite{engstrom2019adversarial, salman2020adversarially} and better approximate the human visual system \cite{schrimpf2020integrative}.
Thus, for crossentropy and entropy regularization, we used an $\epsilon=4$ $L_2$ and $\epsilon=3$ $L_\infty$ robust ResNet50s from \cite{robustness}.
For discriminator regularization, we use the BigGAN class-conditional discriminator with a uniform class vector input (as opposed to a one-hot vector).
For patch adversaries, we train under colorjitter, Gaussian blur, Gaussian noise, random rotation, and random perspective transformations to simulate real-world variations in conditions. 
For region and generalized patch ones, we only use Gaussian blurring and horizontal flipping.
Also for region and generalized patch adversaries, we promote subtlety by penalizing the difference from the original image using the LPIPs perceptual distance \cite{zhang2018unreasonable, lpipstorch}.
All experiments were implemented with PyTorch \cite{NEURIPS2019_9015}, and implementations of all of our work can be run from Google Colab notebooks which we provide. 
We estimate that our project involved a total of < 400 hours of compute, mostly on 12GB NVIDIA Tesla K80 GPUs.

\subsection{How Successful are our Disguises?}
\label{app:how_successful_are_our_disguises}

All of our attacks are designed to be ``adversarial'' in either one or two different ways. First, all are limited to manipulating only a certain portion of the image or latent. 
Second, some of our attacks are trained to be disguised as discussed in Section \ref{sec:methods}.
We subjectively find that our methods for disguise are generally useful. 
For example, compare Fig. \ref{fig:all_printable} to Fig. \ref{fig:only_gen_printable}.
However, some of these features are not well disguised. 
Consider the patch at the top left of Fig. \ref{fig:all_printable} whose target class is a pufferfish but which is disguised as a crane. 

We find it clear that the patch depicts a crane, and the network agrees when shown the patch as a full image.
However, the image also seems to have patterning and coloring that resembles a pufferfish.
Furthermore, when shrunk to the size of a patch and inserted into a source image for an attack, the finer, more crane-like features may be less prominent, and the pufferfish-like ones may be more prominent. 
This suggests that the adversary may be exploiting size biases. 

Ultimately, to the extent that the patch resembles the target class, it is not disguised and arguably not adversarial. 
However, we emphasize three things.
First, we find the disguise classes to generally be much more prominent in these patches than the target class.
Second, being able to recognize target class features when the target class is known versus unknown are very different tasks.
It is key to note such framing effects \cite{hullman2011visualization}.
And third, very few attack methods are always successful at the ImageNet scale, and some need for screening should be expected when optimizing an adversarial example for a complex objective (e.g. universality $+$ transformation robustness $+$ disguise).

We believe the studying human responses to feature-level adversaries and the links between interpretable representations, robustness, and similarity to human cortex \cite{dapello2020simulating} may be promising directions for better understanding both networks and biological brains.

\begin{figure*}
\centering
\includegraphics[width=\linewidth]{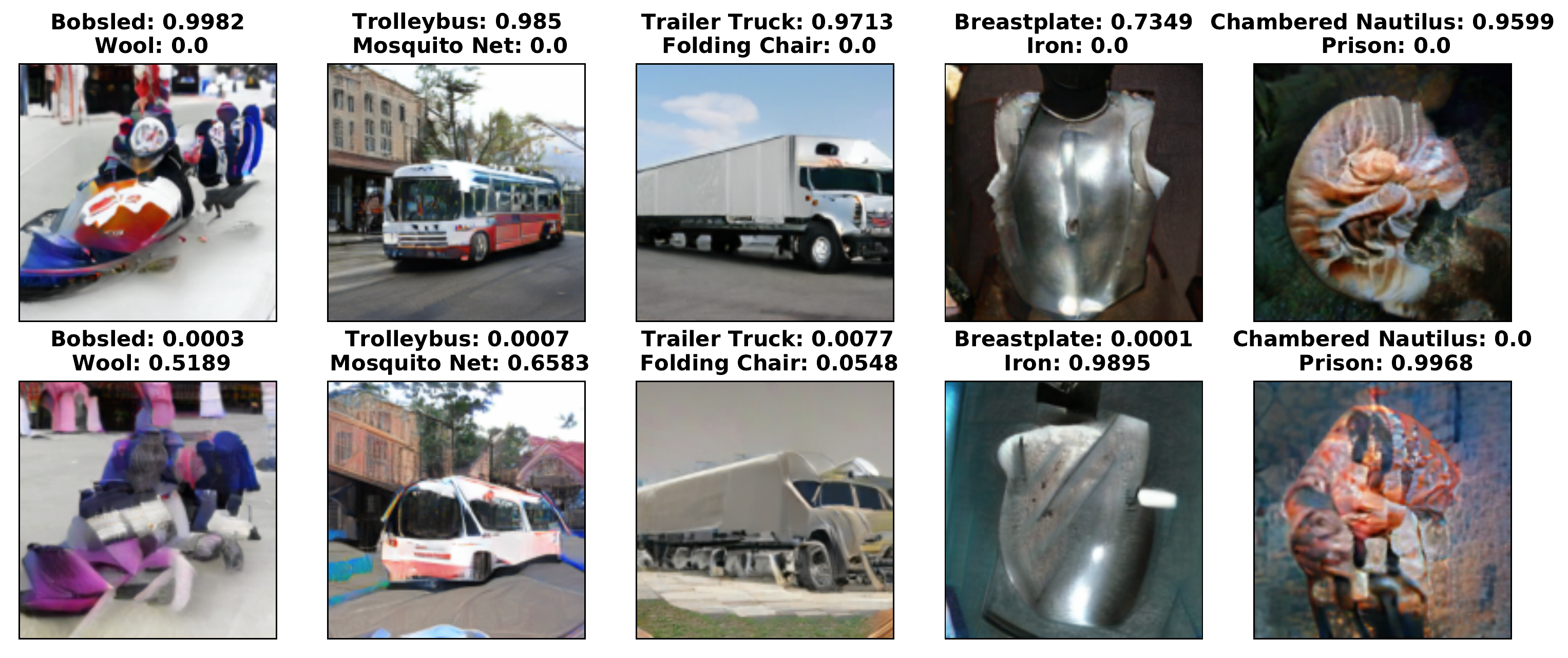}
\caption{Examples of original images (top) alongside class-universal channel adversaries (bottom). Each image is labeled with the source and target class confidence.}
\label{fig:channel}
\end{figure*}

\subsection{Attack Performance versus Disguise} \label{app:fooling_v_disguise}

\begin{figure}
\centering
\includegraphics[width=0.5\linewidth]{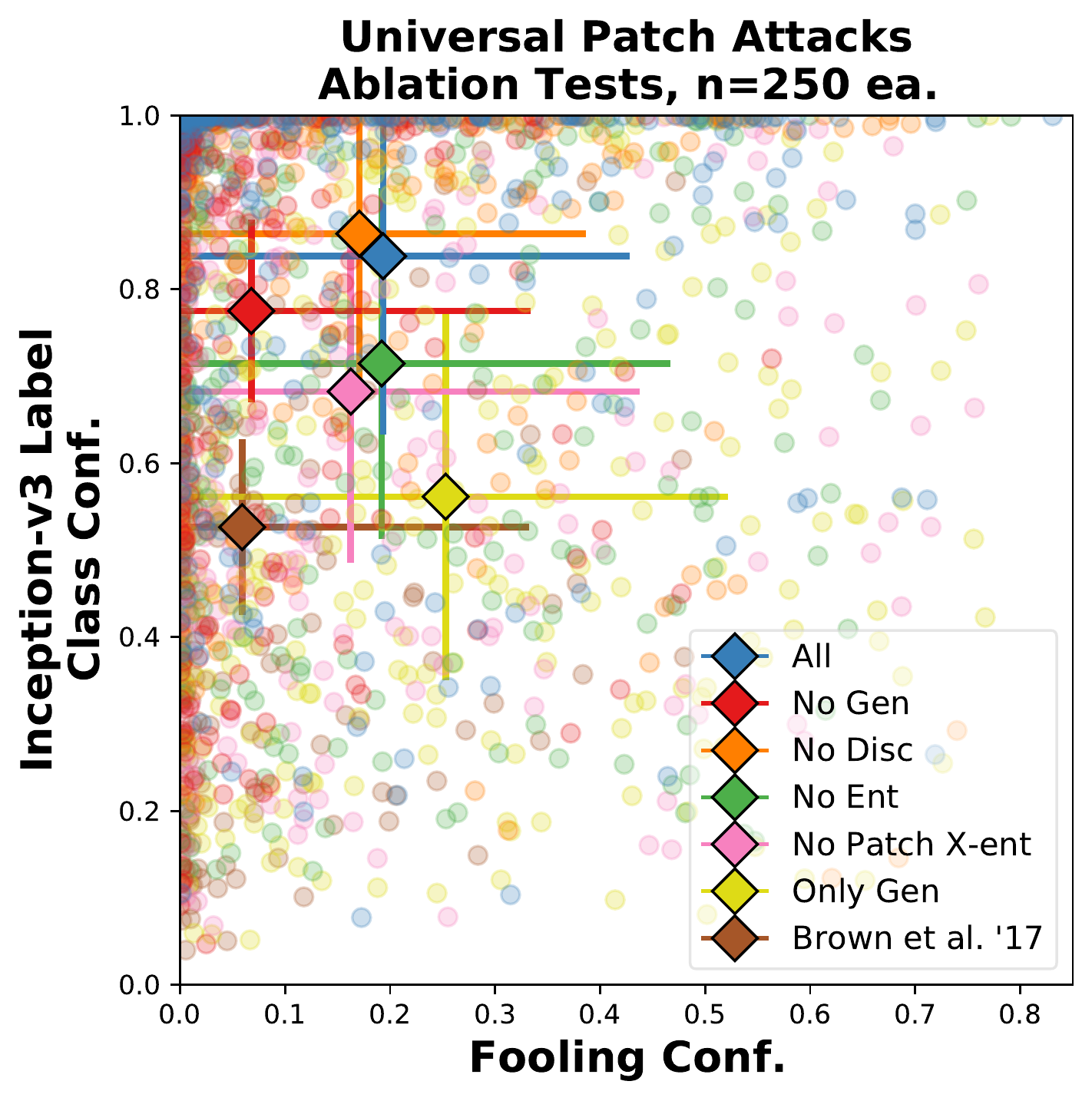}\\
\caption{Targeted, universal patch attacks compared by mean target class confidence and Inception-v3 label-class confidence. Inception-v3 class conf. on the $x$-axis gives the mean target class confidence from the attacked network for images that have the patch inserted. The Inception-v3's label class confidence for the patch on the $y$-axis is used as a proxy for human interpretability. Attacks further up and right are better. Centroids are shown with error bars giving the standard deviation.
}
\label{fig:fooling_v_interpretability}
\end{figure}

In the main paper Section \ref{sec:robust_attacks}, Fig. \ref{fig:violins} plots the successful disguise rate of attacks alongside their distribution of mean target class confidences. 
However, this leaves out how effective each type of attack is at appearing realistic to a human. 
Here, we use the target class confidence of an Inception-v3 \cite{szegedy2016rethinking} as a proxy for how realistic a patch appears to a human. 
Fig. \ref{fig:fooling_v_interpretability} plots the mean target class confidences for successfully-disguised attacks versus their Inception-v3 disguise label confidence. 
This suggests that the attacks that are the best at producing realistic-looking patches are the ``All'' ones with the generator and all regularization terms and the ``No Disc'' ablations which omit the discriminator regularization term.

\subsection{Channel Attacks} \label{app:channel_attacks}

In contrast to the region attacks presented in the main paper, we experiment here with \emph{channel} attacks. 
For region attacks, we optimize an insertion to the latent activations of a generator's layer which spans the channel dimension but not the height and width. 
This is analogous to a patch attack in pixel-space. 
For channel attacks, we optimize an insertion that spans the height and width dimensions but only involves a certain proportion of the channels. 
This is analogous to an attack that only modifies the R, G, or B channel of an image in pixel space. 
Unlike the attacks in Section, \ref{sec:experiments}, we found that it was difficult to create universal channel attacks (single-image attacks, however, were very easy).
Instead, we relaxed this goal and created class-universal ones which are meant to cause any generated example from a source class to be misclassified as a target. 
We also manipulate $1/4$\textsuperscript{th} of the latent instead of $1/8$\textsuperscript{th} as we do for region attacks.
Mean target class confidences and examples from the top 5 attacks out of 16 are shown in Fig. \ref{fig:channel}.
They induce textural changes somewhat like adversaries crafted by \cite{geirhos2018imagenet} and \cite{bhattad2019unrestricted}.

\subsection{Black-Box Attacks} \label{app:black_box_attacks}

\begin{figure*}[h!]
\centering
\includegraphics[width=\linewidth]{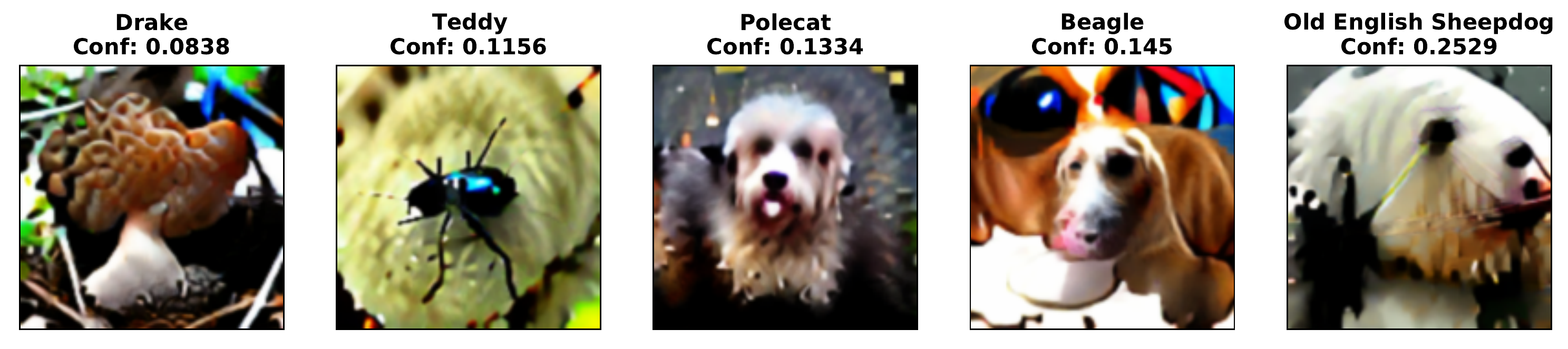}
\includegraphics[width=\linewidth]{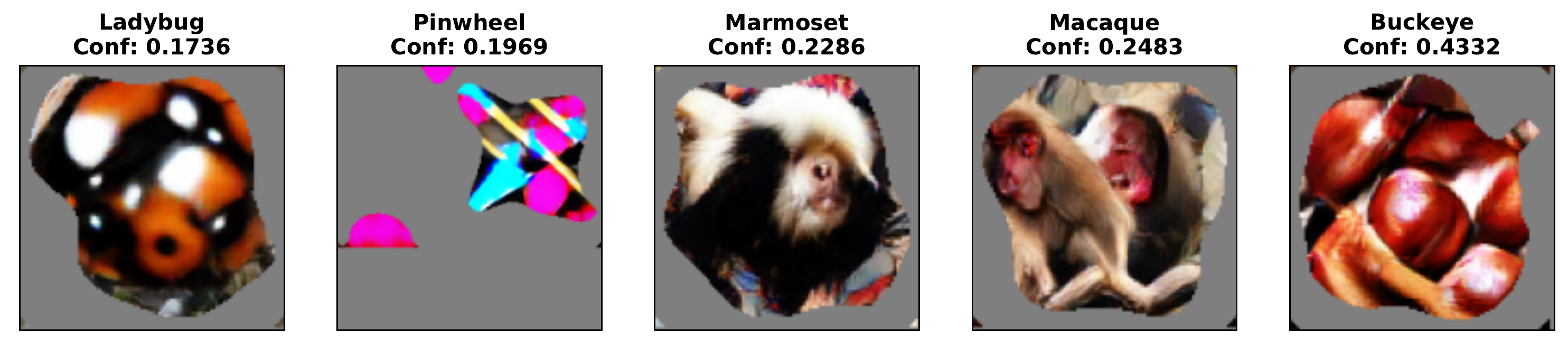}
\caption{Universal black-box adversarial patches (top) and generalized patches (bottom) created using transfer from an ensemble. Patches are displayed alongside their target class and mean target class confidence. }
\label{fig:black_box}
\end{figure*}

Adversaries are typically created using first-order optimization on an input to a network which requires that the parameters are known. 
However, they are often transferrable between models \cite{papernot2016transferability}, and one method for developing black-box attacks is to train against a different model and then transfer to the intended target.
We do this for our adversarial patches and generalized patches by attacking a large ensemble of AlexNet \cite{krizhevsky2012imagenet}, VGG19 \cite{simonyan2014very}, Inception-v3 \cite{szegedy2016rethinking}, DenseNet121 \cite{huang2017densely}, ViT\cite{dosovitskiy2020image, vit}, and two robust ResNet50s \cite{robustness}, and then transferring to ResNet50 \cite{he2016deep}. 
Otherwise, these attacks were identical to the ones in Fig. \ref{fig:examples} including a random source/target class and optimization for disguise. 
Many were unsuccessful, but a sizable fraction were able to work on the ResNet50 with a mean confidence of over 0.1 for randomly sampled images. 
The top 5 out of 64 of these attacks for patch and generalized patch adversaries are shown in Fig. \ref{fig:black_box}. 

\subsection{Discovering Feature-Class Associations} \label{app:discovering_feature_class_associations}

\begin{figure*}[h!]
\centering
\begin{tabular}{cc}
    \includegraphics[width=0.45\linewidth]{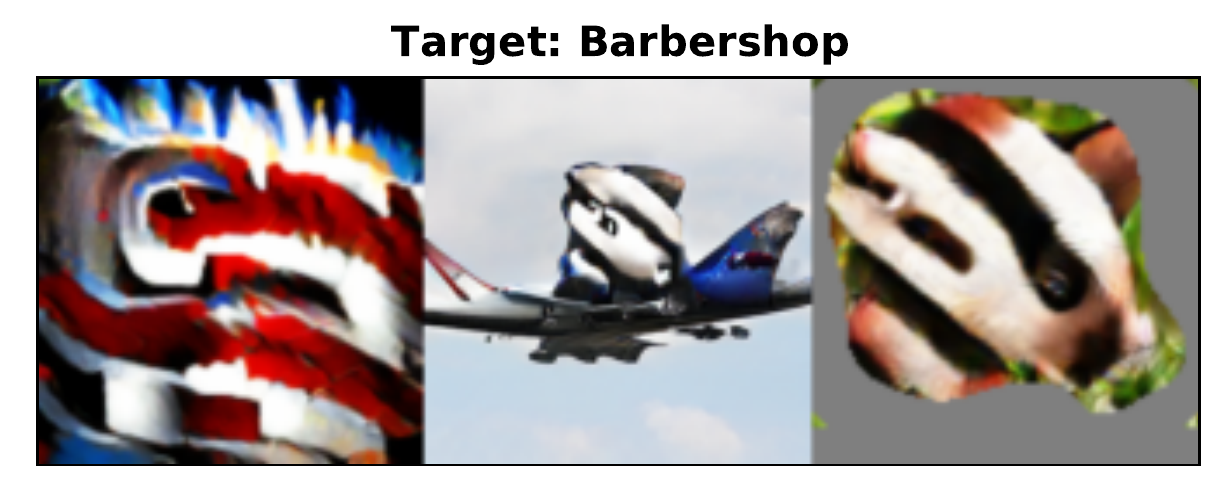} &
	\includegraphics[width=0.45\linewidth]{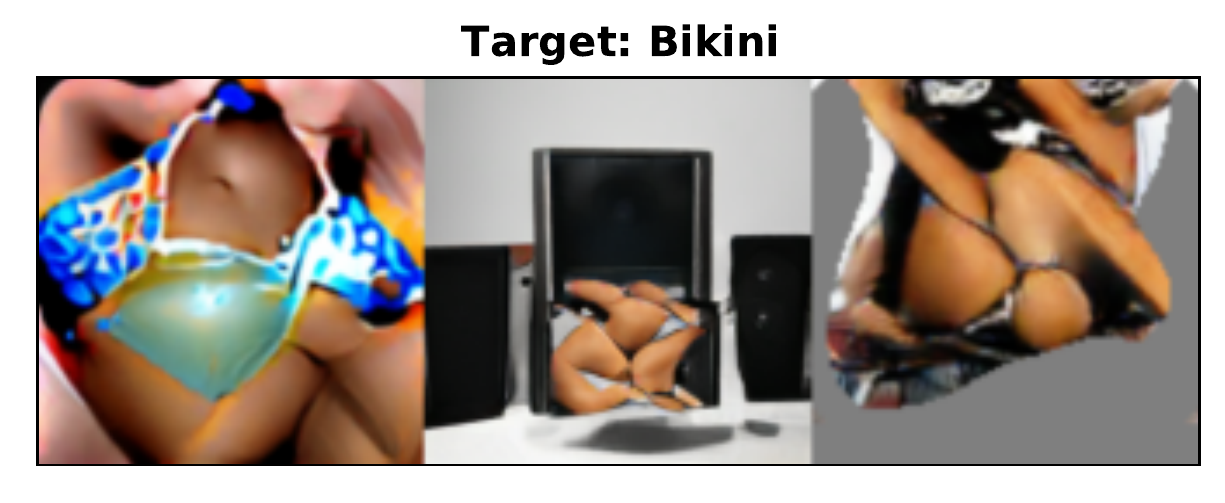}
\end{tabular}
\caption{Examples of good and bad features class associations in which barber pole stripes are associated with a barbershop and a particular type of skin is associated with a bikini. Patch (left), region (middle), and generalized patch adversaries (right) are shown.}
\label{fig:barbershop_bikini}
\end{figure*}

Fig. \ref{fig:barbershop_bikini} shows two simple examples of using feature-level attacks to identify feature-class associations. It shows one positive example in which the barbershop class is desirably associated with barber-pole-stripe-like features and one negative example in which the bikini class is undesirably associated with a particular type of skin. 
Notably, the ability to simply identify features associated with a target class is not a unique capability of our attacks and could also be achieved with feature visualization \cite{olah2017feature}.

\begin{figure*}[h!]
\centering
\includegraphics[width=\linewidth]{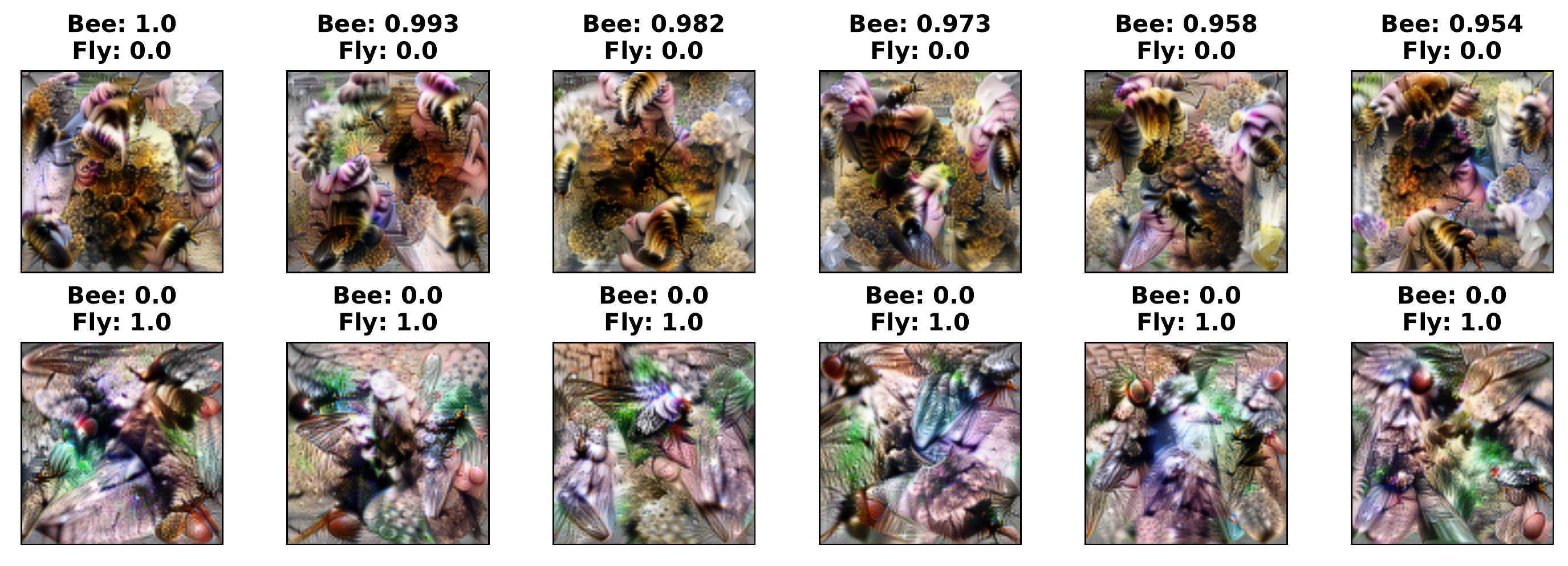}\\
\bigskip
\includegraphics[width=\linewidth]{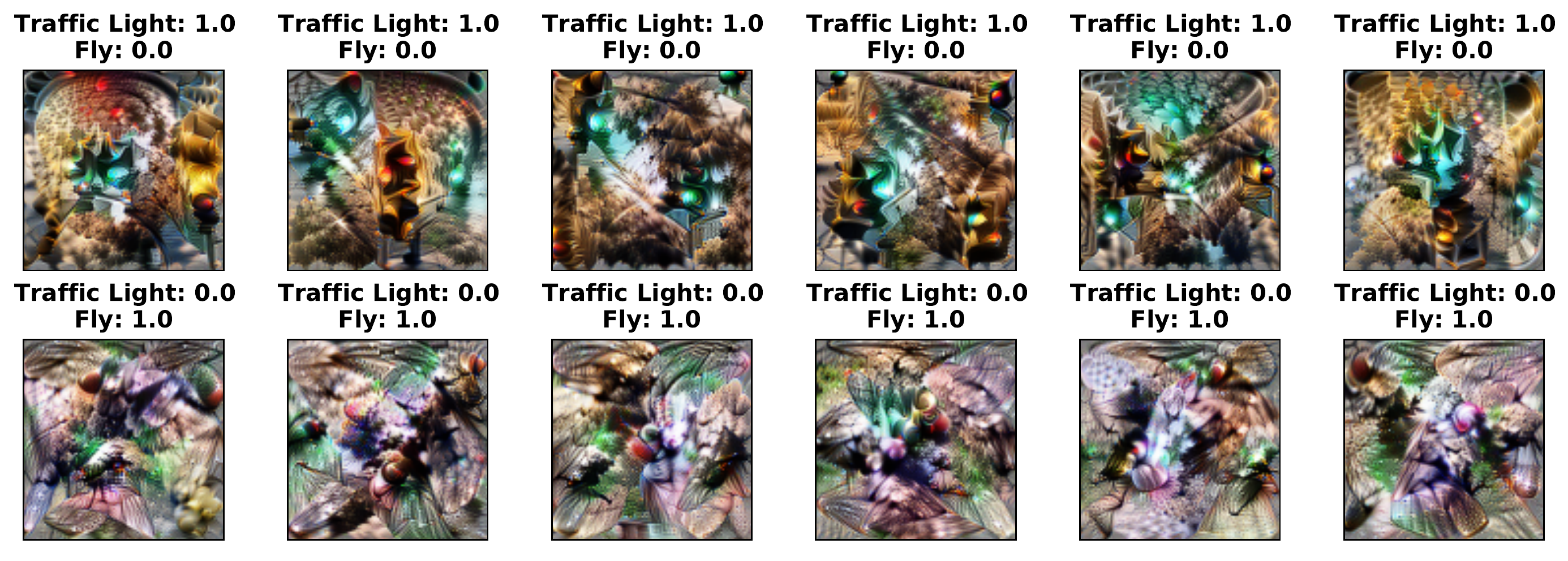}\\
\bigskip
\includegraphics[width=\linewidth]{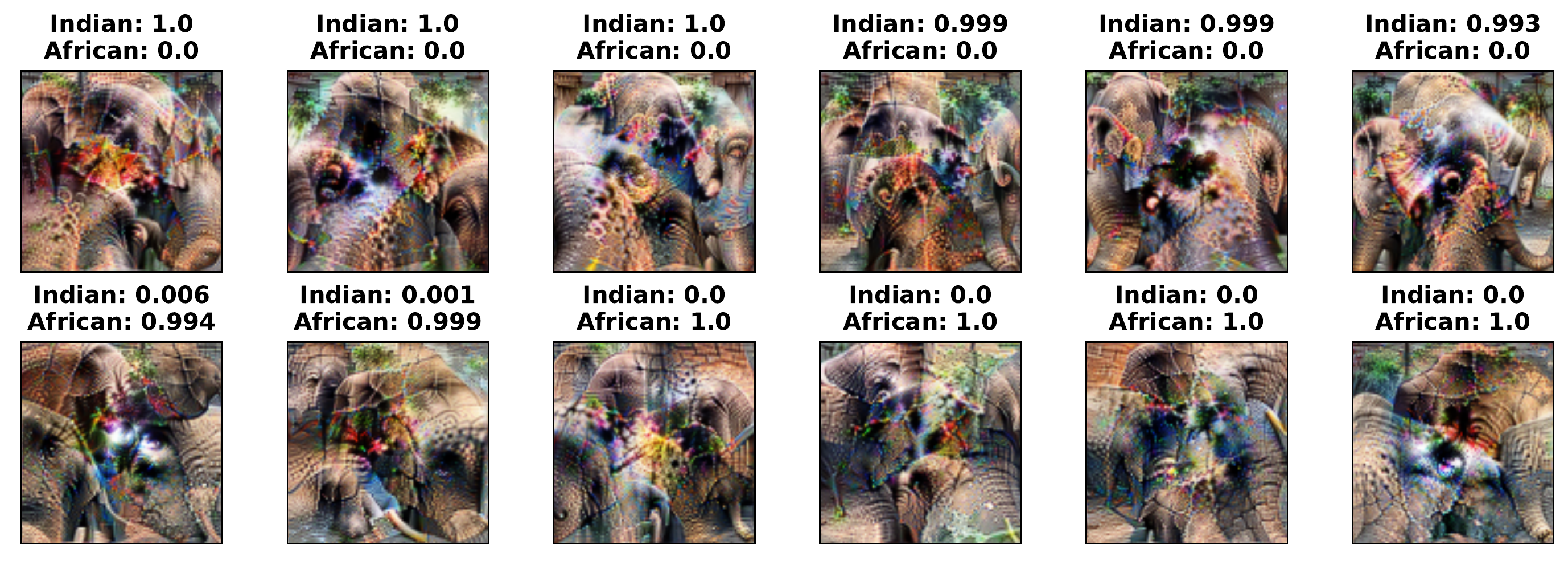}\\
\bigskip
\includegraphics[width=\linewidth]{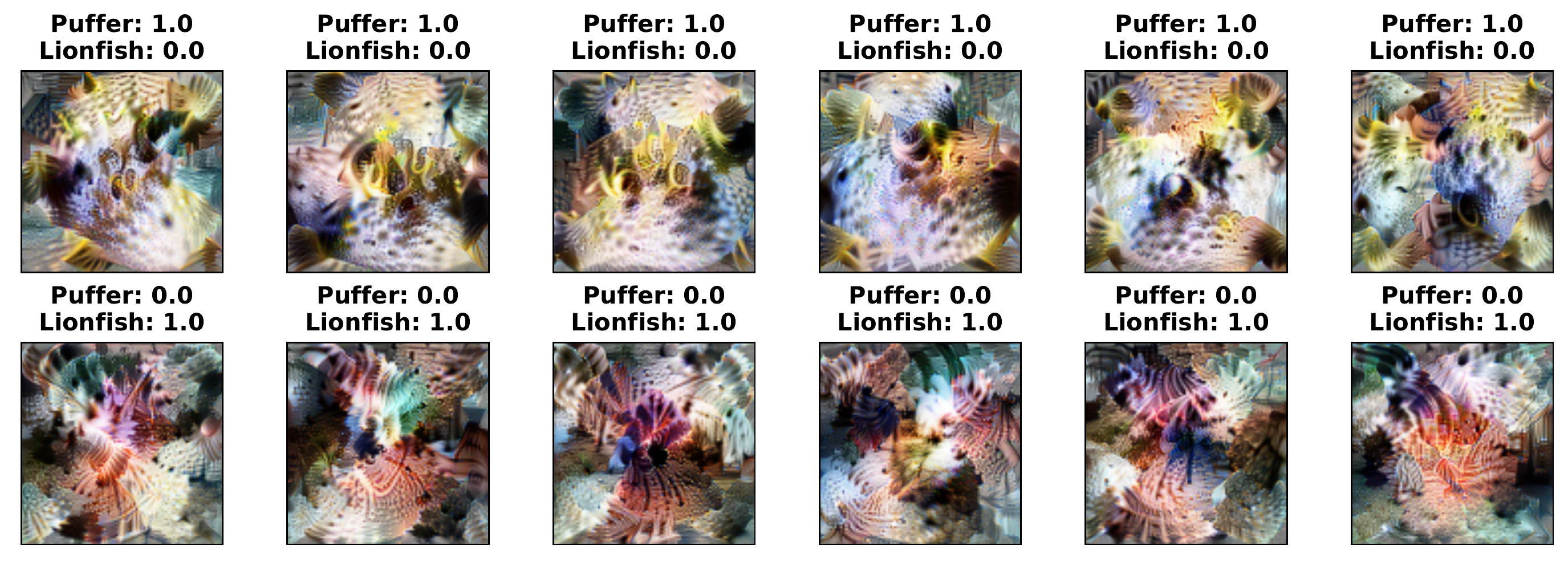}
\caption{Class impressions for four pairs of classes. These could be used for providing insight about copy/paste attacks in a similar way to the examples from Fig. \ref{fig:copy_paste}. Each subfigure is labeled with the network's output confidence for both classes.}
\label{fig:class_impressions}
\end{figure*}

\subsection{Copy/Paste Attacks: Feature-Level Adversaries vs. Class Impressions} \label{app:copy_paste_attacks_with_class_impressions}

\cite{mu2020compositional, hernandez2022natural, carter2019activation} each used interpretability methods to guide the development of copy/paste adversaries. 
\cite{mu2020compositional} and \cite{hernandez2022natural}, used network dissection \cite{bau2017network} to develop interpretations of neurons and fit a semantic description to neurons which they combined with analysis of weight magnitudes.
This allowed them to identify cases in which the networks learned undesirable feature-class associations.
However, this approach cannot be used to make a targeted search for copy/paste attacks that will cause a given source class to be misclassified as a given target. 

More similar to our work is 
\cite{carter2019activation} who found inspiration for successful copy/paste adversaries by creating a dataset of visual features and comparing the differences between ones which the network assigned the source versus target label.
We take inspiration from this approach to create a baseline against which to compare our method for designing copy/paste attacks from Section \ref{sec:interpretable_copy_paste_attacks}.
Given a source and target class such as a bee and a fly, we optimize a set of inputs to a network for each class in order to maximize the activation of the output node for that class. 
\cite{mopuri2018ask} refers to these as \emph{class impressions}.
We develop these inputs using the Lucent \cite{lucent} package.
We do this for the same source/target class pairs as in Fig. \ref{fig:copy_paste} and display 6 per class in Fig. \ref{fig:class_impressions}.
In each subfigure, the top row gives class impressions of the source class, and the bottom gives them for the target.
Each class impression is labeled with the network's confidences for the source and target class. 
In analyzing these images, we find limited evidence of traffic-light-like features in the fly class impressions, and we find no evidence of more blue coloration in the African elephant class impressions than the Indian ones. 

These class impressions seem comparable but nonredundant with our method from Section \ref{sec:interpretable_copy_paste_attacks}.
However, our approach may have an advantage over the use of class impressions in that it is equipped to design features that look like the target class \emph{conditional} on a distribution of source images.
In contrast, a class impression is only meant to visualize features typical of the target class. 
This may be why our adversarial attacks are able to suggest that inserting a traffic light into a bee image or a blue object into an Indian elephant image can cause a misclassification as a fly or African elephant respectively. 
Bees and Flies and Indian and African elephants are pairs of similar classes.
For cases in which the source and target are related, the class impressions may share many of the same features and thus be less effective for identifying adversarial features that work \emph{conditional} on the source image distribution. 

\subsection{Defense via Adversarial Training} \label{app:defense_via_adversarial_training}

Adversarial training is a common effective means for improving robustness.
Here, to test how effective it is for our attacks, for 5 pairs of similar classes, we generate datasets of 1024 images evenly split between each class and between images with and without adversarial perturbations.
This prevents the network from learning to make classifications based on the mere presence or absence of a patch.
We do this separately for channel, region, and patch adversaries before treating the victim network as a binary classifier and training on the examples. 
We report the post-training minus pre-training accuracies in Tbl. \ref{tbl:adv_training} and find that across the class pairs and attack methods, the adversarial training improves binary classification accuracy by a mean of 42\%.

\begin{table}[h!]
\centering
\begin{tabular}{|l|l|l|l|l|}
\hline
                                & \textbf{Channel} & \textbf{Region} & \textbf{Patch} & \textbf{Mean} \\ \hline
\textbf{Great White/Grey Whale} & 0.49             & 0.29            & 0.38           & 0.39          \\ \hline
\textbf{Alligator/Crocodile}    & 0.13             & 0.29            & 0.60           & 0.34          \\ \hline
\textbf{Lion/Tiger}             & 0.29             & 0.28            & 0.63           & 0.40          \\ \hline
\textbf{Frying Pan/Wok}         & 0.32             & 0.39            & 0.68           & 0.47          \\ \hline
\textbf{Scuba Diver/Snorkel}    & 0.42             & 0.36            & 0.69           & 0.49          \\ \hline
\textbf{Mean}                   & 0.33             & 0.32            & 0.60           & \textbf{0.42} \\ \hline
\end{tabular}
\vspace{0.5cm}
\caption{Binary classification accuracy improvements from adversarial training for channel, region, and patch adversaries across 5 class pairs.}
\label{tbl:adv_training}
\end{table}

\subsection{Examples: Resizable, Printable Patches} \label{app:printable_examples}

\begin{figure*}[p!]
\centering
\includegraphics[width=0.55\linewidth]{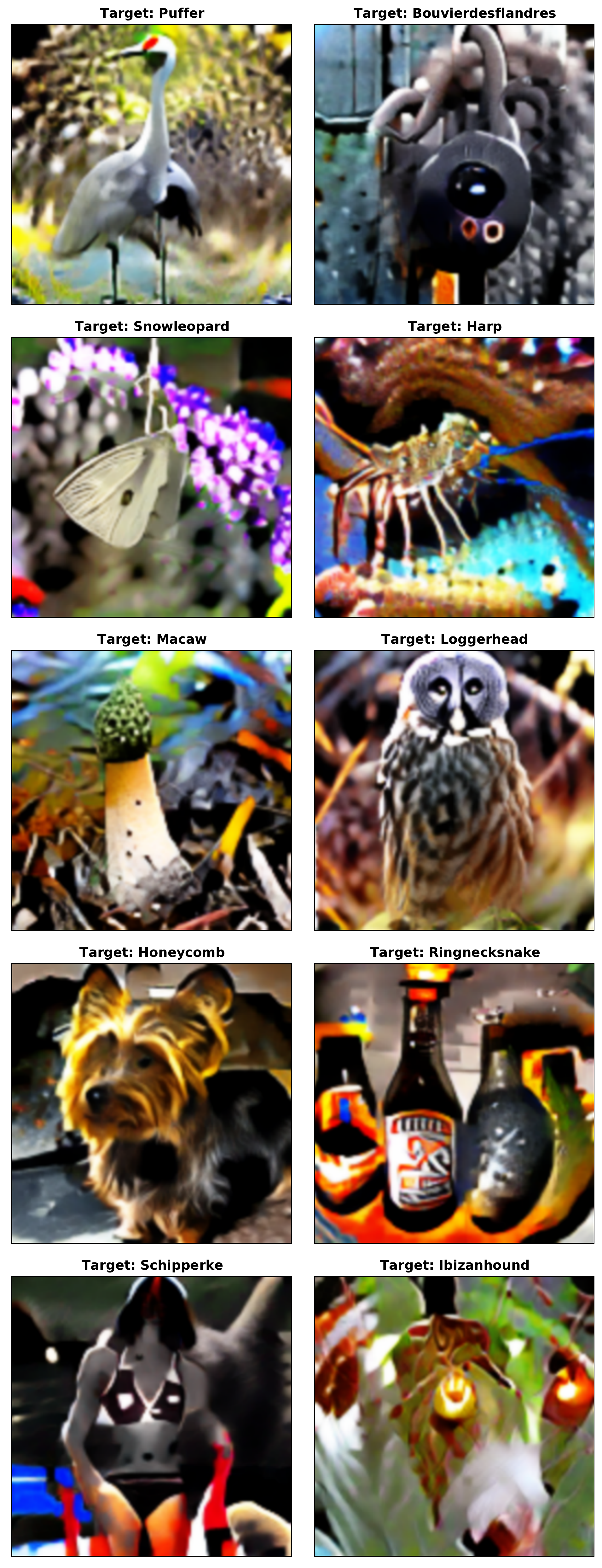}
\caption{Examples of attacks created with a generator and with the disguise regularization terms from Eq. \ref{eq:l_reg}. See section \ref{sec:robust_attacks} and the ``All'' datapoints from Fig. \ref{fig:violins}. We printed these to test physical realizability in Section \ref{sec:robust_attacks}}
\label{fig:all_printable}
\end{figure*}

\begin{figure*}[p!]
\centering
\includegraphics[width=0.55\linewidth]{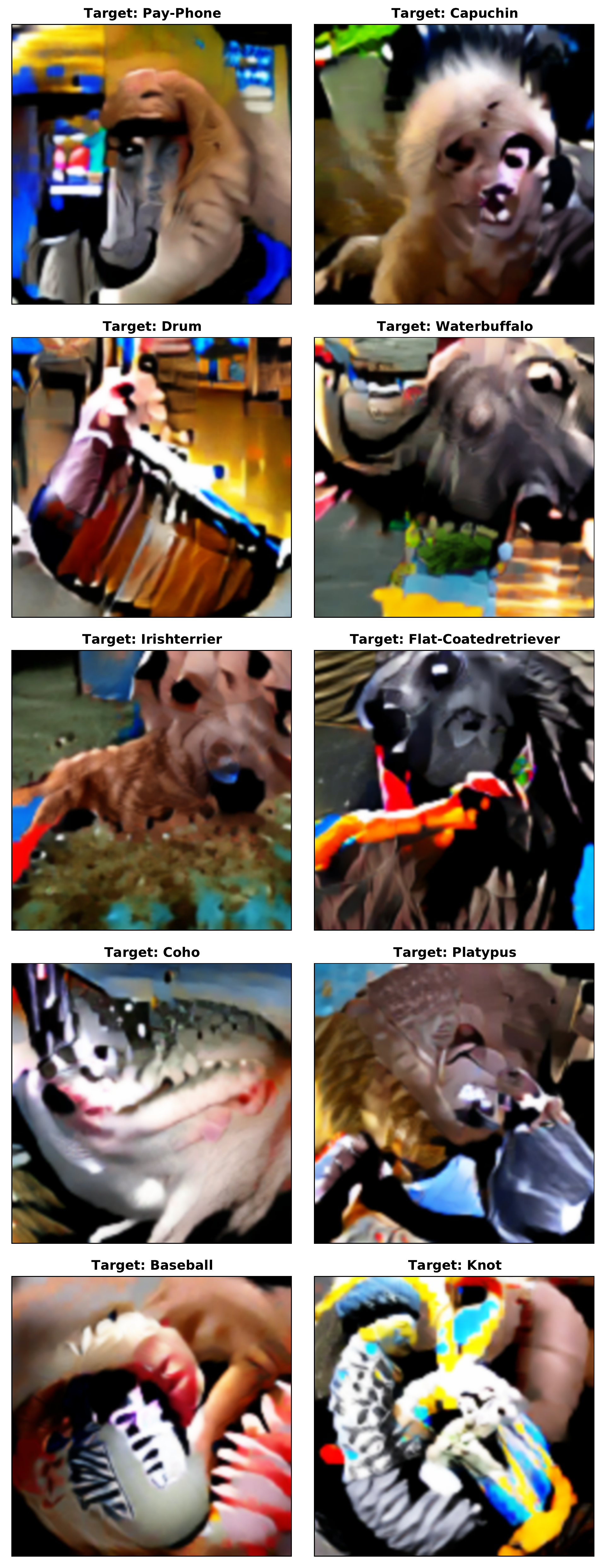}
\caption{Examples of attacks created with a generator. See section \ref{sec:robust_attacks} and the ``Only Gen'' datapoints from Fig. \ref{fig:violins}.}
\label{fig:only_gen_printable}
\end{figure*}

\begin{figure*}[p!]
\centering
\includegraphics[width=0.6\linewidth]{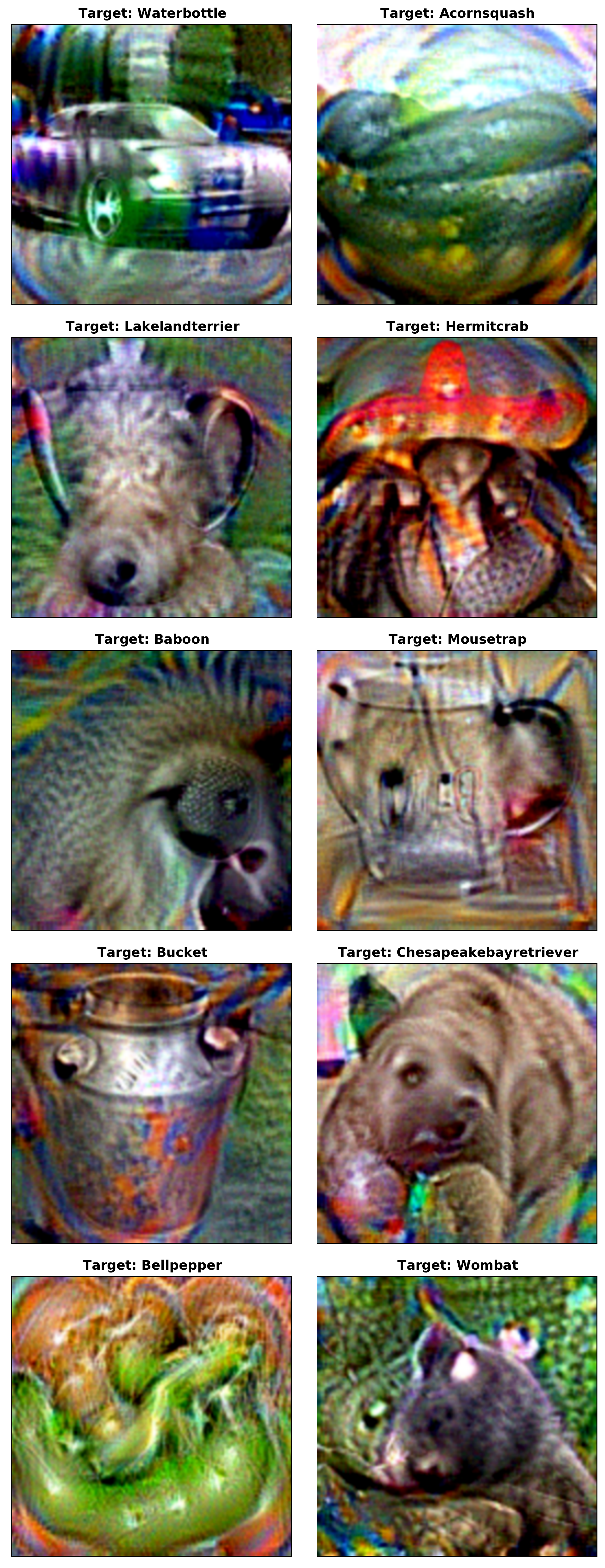}
\caption{Examples of pixel-space attacks \cite{brown2017adversarial}. See section \ref{sec:robust_attacks} and the ``Brown et al. '17'' datapoints from Fig. \ref{fig:violins}.}
\label{fig:brown_printable}
\end{figure*}

See Figs. \ref{fig:all_printable}, \ref{fig:only_gen_printable}, and \ref{fig:brown_printable} for feature-level and pixel-level control adversarial images.
We encourage readers to experiment with these images (which were optimized to attack a ResNet50) or with others that can be created using our provided code.
In doing so, one might find a mobile app to be convenient. 
We used Photo Classifier.\footnote{https://apps.apple.com/us/app/photo-classifier/id1296922017} 

\subsection{High-Level Summary} \label{app:jargon_free_summary}

Here, we provide an easily readable summary of this work that avoids jargon. 

\medskip

Historically, it has proven difficult to write conventional computer programs that classify images. 
But recently, immense progress has been made through neural networks which can now classify images into hundreds or thousands of categories with high accuracy. 
Despite this performance, we still don't fully understand the features that they use to classify images. 
Somewhat worryingly, past work has demonstrated that it is easy to take an image that the network classifies correctly and perturb its pixel values by a tiny amount in such a way that the network will misclassify it. 
For example, we can take a cat, make human-imperceptible changes to the pixels, and make the network believe that it is a dog. 
This process of designing an image that the network will misclassify is called an ``adversarial attack.''

Unfortunately, conventional adversarial attacks tend to produce perturbations that are not interpretable.
To a human, they usually appear as pixelated noise (when exaggerated to be visible). 
As a result, they do not help us understand how networks will process human-describable inputs, and they do little to help us understand practical flaws in networks. 
Here, our goal is to develop different types of adversarial features that can be useful as debugging tools for these networks.

There are two properties that we want our adversarial images to have. 
First, we want them to be feature-level (not pixel-level) so that a human can interpret
them, and second, we want them to be robust so that interpretations are generalizable. 
In one sense, this is not a new idea. 
Quite the opposite, in fact -- there are examples of this in the animal kingdom.
Figure \ref{fig:nature} shows examples of adversarial eyespots on a peacock and butterfly and adversarial patterns on a mimic octopus. 
These are interpretable, robust to a variety of viewing conditions, and give us useful information about how biological brains process these visual features. 

The key technique that we use to create adversarial images involves an image-generating network. 
Instead of perturbing pixels to cause an image to be misclassified, we perturb the internal state of the generator in order to induce a feature-level change to the output image. 
We also found that optimizing adversarial features under transformations to the images (like blurring, cropping, rotating, etc.) and using some additional terms in our optimization objective to encourage more realistic and better-disguised images improved results. 

Our first key finding is that these attacks are very robust and versatile. 
For example, among other capabilities, we demonstrate that they can be universal (working for any image to which we apply them) and physically-realizable (working in the physical world when printed and photographed).

Our second key finding is that these adversaries are useful for interpreting networks. 
We use them as a way of producing useful classes of inputs for understanding ways that they can fail. 
Our strategy is to prove that these adversaries can help us understand the network well enough to exploit it. 
We analyze our adversarial features to get ideas for real-world features that resemble them. 
Then we test this interpretation by creating ``copy/paste'' attacks in which one natural image is pasted into another in order to cause a particular misclassification. 
Some of these are unexpected. 
For example, in Fig. \ref{fig:copy_paste}, we find that a traffic light can make a bee look like a fly.

Together, our findings suggest that feature-level adversaries are very versatile attacks and practical debugging tools for finding flaws in networks. 
One implication is that by training networks on these adversaries, we might be able to make them more robust to failures that are due to feature-level properties of images. 
We also argue that these adversaries should be used as a practical debugging tool to diagnose problems in networks.

\end{document}